%% file: nonconvex-arxiv.tex
\documentclass[11pt]{article} 
\usepackage{iclr2017_conference,times}
\iclrfinalcopy
\usepackage{hyperref}
\usepackage{url}
\usepackage{caption}
\usepackage{subcaption}

\usepackage{amsmath,graphicx}
\usepackage{algorithm}
\usepackage[noend]{algpseudocode}

\makeatletter
\def\BState{\State\hskip-\ALG@thistlm}
\makeatother

\usepackage{amssymb}
\usepackage{amsthm}
\usepackage{geometry}
\usepackage{graphicx}
\usepackage{epstopdf}
\usepackage{hyperref}
\usepackage[applemac]{inputenc}
\usepackage{color}

\newcommand {\R} {{\mathbb{R}}}

\newcommand {\E} {{\mathbb{E}}}

\newcommand {\Fem} {F_{e}}
\newcommand {\Forr} {F_{o}}

\usepackage{amsthm}
\usepackage{amsmath}
\usepackage{amssymb}

\newtheorem{theorem}{Theorem}[section]

\newtheorem{proposition}[theorem]{Proposition}
\newtheorem{corollary}[theorem]{Corollary}

\title{Topology and Geometry of Half-Rectified Network Optimization }

\author{C. Daniel Freeman \\
Department of Physics, \\
UC Berkeley, Berkeley, CA 94720, USA \\
\texttt{daniel.freeman@berkeley.edu} \\ 
Joan Bruna \thanks{Currently on leave from UC Berkeley.} \\
Courant Institute of Mathematical Sciences \\ 
New York University, New York, NY 10011, USA \\
\texttt{bruna@cims.nyu.edu} \\ 
}

\begin{document}

\maketitle

\begin{abstract}
The loss surface of deep neural networks has recently attracted interest 
in the optimization and machine learning communities as a prime example of 
high-dimensional non-convex problem. Some insights were recently gained using spin glass 
models and mean-field approximations, but at the expense of strongly simplifying the nonlinear nature of the model.

In this work, we do not make any such assumption and study conditions 
on the data distribution and model architecture that prevent the existence 
of bad local minima. Our theoretical work quantifies and formalizes two 
important \emph{folklore} facts: (i) the landscape of deep linear networks has a radically different topology 
from that of deep half-rectified ones, and (ii) that the energy landscape 
in the non-linear case is fundamentally controlled by the interplay between the smoothness of the data distribution and model over-parametrization. Our main theoretical contribution is to prove that half-rectified single layer networks are asymptotically connected, and we provide explicit bounds that reveal the aforementioned interplay.

The conditioning of gradient descent is the next challenge we address. 
We study this question through the geometry of the level sets, and we introduce
an algorithm to efficiently estimate the regularity of such sets on large-scale networks. 
Our empirical results show that these level sets remain connected throughout 
all the learning phase, suggesting a near convex behavior, but they become 
exponentially more curvy as the energy level decays, in accordance to what is observed in practice with 
very low curvature attractors.
\end{abstract}

\input{intro}

\input{topology}

\input{geometry}

\input{experiments}

\input{conclusions}


\subsubsection*{Acknowledgments}

We would like to thank Mark Tygert for pointing out 
the reference to the $\epsilon$-nets and Kolmogorov capacity, and Martin Arjovsky for spotting 
several bugs in early version of the results. 
 We would also like to thank Maithra Raghu and Jascha Sohl-Dickstein for enlightening discussions, as well as Yasaman Bahri for helpful feedback on an early version of the manuscript.  CDF was supported by the NSF Graduate Research Fellowship under Grant DGE-1106400.

\bibliographystyle{abbrv} 
\bibliography{iclr2017_conference}

\appendix

\input{constrained}

\input{proofs}

\input{visualization}

\end{document}

%% file: intro.tex
\section{Introduction}
\label{sec:Intro}

%

Optimization is a critical component in deep learning, governing its success in different areas of computer vision, speech processing and natural language processing. The prevalent optimization strategy is Stochastic Gradient Descent, invented by Robbins and Munro in the 50s. The empirical performance of SGD on these models is better than one could expect in generic, arbitrary non-convex loss surfaces, often aided by modifications yielding significant speedups \cite{duchi2011adaptive, hinton2012lecture, ioffe2015batch, kingma2014adam}. This raises a number of theoretical questions as to why neural network optimization does not suffer in practice from poor local minima. 

The loss surface of deep neural networks has recently attracted interest 
in the optimization and machine learning communities as a paradigmatic example of 
a hard, high-dimensional, non-convex problem. 
Recent work has explored models from statistical physics such as spin glasses \cite{choromanska2015loss}, 
in order to understand the macroscopic properties of the system, but at the expense of strongly simplifying the nonlinear nature of the model. Other authors have advocated 
that the real danger in high-dimensional setups are saddle points 
rather than poor local minima \cite{dauphin2014identifying}, although 
recent results rigorously establish that gradient descent does not 
get stuck on saddle points \cite{lee2016gradient} but merely slowed down. 
Other notable recent contributions are \cite{kawaguchi2016deep}, which further develops the spin-glass 
connection from \cite{choromanska2015loss} and resolves the linear case by showing that no poor local minima exist; \cite{sagun2014explorations} which also discusses the impact of stochastic vs plain gradient,  \cite{soudry2016no}, that studies Empirical Risk Minimization for piecewise multilayer neural networks under overparametrization (which needs to grow with the amount of available data),  
and \cite{goodfellow2014qualitatively}, which provided insightful intuitions on the loss surface of large deep learning models and partly motivated our work. Additionally, the work \cite{safran2015quality} studies some topological 
properties of homogeneous nonlinear networks and shows how overparametrization acts upon these properties, and the pioneering \cite{shamir2} studied the distribution-specific hardness of optimizing non-convex objectives.  Lastly, several papers submitted concurrently and independently of this one deserve note, particularly \cite{Swirszcz2016} which analyzes the explicit criteria under which sigmoid-based neural networks become trapped by poor local minima, as well as \cite{Tian2017}, which offers a complementary study of two layer ReLU based networks, and their learning dynamics.

In this work, we do not make any linearity assumption and study conditions 
on the data distribution and model architecture that prevent the existence 
of bad local minima. 
The loss surface $F(\theta)$ of a given model can be expressed in terms of its level sets $\Omega_\lambda$, which contain for each energy level $\lambda$ all parameters $\theta$ yielding a loss smaller or equal than $\lambda$. A first question we address concerns the topology of these level sets, i.e. under which conditions they are connected. Connected level sets imply that one can always find a descent direction at each energy level, and therefore that no poor local minima can exist. In absence of nonlinearities, deep (linear) networks have connected level sets \cite{kawaguchi2016deep}. We first generalize this result to include ridge regression (in the two layer case) and provide an alternative, more direct proof of the general case. We then move to the half-rectified case and show that the topology is intrinsically different and clearly dependent on the interplay between data distribution and model architecture. Our main theoretical contribution is to prove that half-rectified single layer networks are asymptotically connected, and we provide explicit bounds that reveal the aforementioned interplay.

Beyond the question of whether the loss contains poor local minima or not, the immediate follow-up question that determines the convergence of algorithms in practice is the local conditioning of the loss surface. It is thus related not to the topology but to the shape or geometry of the level sets. As the energy level decays, one expects the level sets to exhibit more complex irregular structures, which correspond to regions where $F(\theta)$ has small curvature. In order to verify this intuition, we introduce an efficient algorithm to estimate the geometric regularity of these level sets by approximating geodesics of each level set starting at two random boundary points. Our algorithm uses dynamic programming and can be efficiently deployed to study mid-scale CNN architectures on MNIST, CIFAR-10 and RNN models on Penn Treebank next word prediction. 
Our empirical results show that these models have a nearly convex behavior up until their lowest test errors, with a single connected component that becomes more elongated as the energy decays. 
The rest of the paper is structured as follows. Section 2 presents our theoretical results on the topological connectedness of multilayer networks. Section 3 presents our path discovery algorithm and Section 4 covers the numerical experiments.

%% file: topology.tex
\section{Topology of Level Sets}

Let $P$ be a probability measure on a product space $\mathcal{X} \times \mathcal{Y}$, 
where we assume $\mathcal{X}$ and $\mathcal{Y}$ are Euclidean vector spaces for simplicity.
Let $\{ (x_i, y_i)\}_i$ be an iid sample of size $L$ drawn from $P$ defining the training set.
We consider the classic empirical risk minimization of the form
\begin{equation}
\label{emp_risk_min}
\Fem(\theta) = \frac{1}{L} \sum_{l=1}^L \| \Phi(x_i;\theta) - y_i \|^2 + \kappa \mathcal{R}(\theta)~,
\end{equation}
where $\Phi(x ; \theta)$ encapsulates the feature representation 
that uses parameters $\theta \in \R^S$ and $\mathcal{R}(\theta)$ is a regularization term. 
 In a deep neural network, $\theta$
contains the weights and biases used in all layers.
For convenience, in our analysis we will also use the oracle risk minimization:
\begin{equation}
\label{risk_min}
\Forr(\theta) = \E_{(X,Y) \sim P} \| \Phi(X;\theta) - Y \|^2 + \kappa \mathcal{R}(\theta)~.
\end{equation}
Our setup considers the case where $\mathcal{R}$ 
consists on either $\ell_1$ or $\ell_2$ norms, as we shall describe below.
They correspond to well-known sparse and ridge regularization respectively.
%

\subsection{Poor local minima characterization from topological connectedness}

We define the level set of $F(\theta)$ as 
\begin{equation}
\Omega_F(\lambda) = \{ \theta \in \R^S~;~F(\theta) \leq \lambda \}~. 
\end{equation}

The first question we study is the structure of critical points of $\Fem(\theta)$ and $\Forr(\theta)$
when $\Phi$ is a multilayer neural network. 
For simplicity, we consider first a strict notion of local minima: 
$\theta \in \R^S $ is a strict local minima of $F$ if there is $\epsilon>0$ with $F(\theta') > F(\theta)$ for all $\theta' \in B(\theta,\epsilon)$ and $\theta'\neq \theta$.
In particular, we are interested to know whether
$\Fem$ has local minima which are not global minima. 
This question is answered by 
knowing whether $\Omega_F(\lambda)$ is connected at each energy level $\lambda$:

\begin{proposition}
\label{connectedminima}
If $\Omega_F(\lambda)$ is connected for all $\lambda$ then every local minima of $F(\theta)$ is a global minima. 
\end{proposition}

Strict local minima implies that $\nabla F(\theta) =0$ and $H F(\theta) \succeq 0$, but avoids degenerate cases where $F$ is constant along a manifold intersecting $\theta$. In that scenario, if $\mathcal{U}_\theta$ denotes that manifold, our reasoning immediately 
implies that if $\Omega_F(\lambda)$ are connected, then for all $\epsilon > 0$ there exists $\theta'$ with $\text{dist}(\theta',\mathcal{U}_\theta) \leq \epsilon$ and $F(\theta') < F(\theta)$. In other words, some element at the boundary of $\mathcal{U}_\theta$ must be a saddle point. 
A stronger property that eliminates the risk of gradient descent getting stuck at $\mathcal{U}_\theta$ is 
  that \emph{all} elements at the boundary of $\mathcal{U}_\theta$ are saddle points. This can be guaranteed if one can show that 
  there exists a path connecting any $\theta$ to the lowest energy level such that $F$ is strictly decreasing along it. 

Such degenerate cases arise in deep linear networks in absence of regularization. If $\theta = (W_1, \dots, W_K)$ denotes any parameter value, with $N_1, \dots N_K$ denoting the hidden layer sizes, and $F_k \in \mathbf{GL}_{N_k}^{+}(\R)$ are arbitrary elements of the general linear group of invertible $N_k \times N_k$ matrices with positive determinant, then   
$$\mathcal{U}_\theta = \{ W_1 F_1^{-1}, F_1 W_2 F_2^{-1}, \dots, F_K W_K  ~;~ F_k \in \mathbf{GL}_{N_k}^{+}(\R) \}~. $$
In particular, $\mathcal{U}_\theta$ has a Lie Group structure. In the half-rectified nonlinear case, the general linear group is replaced by the Lie group of homogeneous invertible matrices $F_k = \text{diag}(\alpha_1, \dots, \alpha_{N_k})$ with $\alpha_j > 0$. 

This proposition shows that a sufficient condition to prevent the existence of poor local minima is having connected level sets, but this condition is not necessary: one can have isolated local minima lying 
at the same energy level. This can be the case in systems that are defined up to 
a discrete symmetry group, such as multilayer neural networks. However, as we shall see next, this case puts the system in a brittle position, since one needs to be able to account for all the local minima (and there can be exponentially many of them as the parameter dimensionality increases) and verify that their energy is indeed equal. 

\subsection{The Linear Case}

We first consider the 
particularly simple case where 
$F$ is a multilayer network defined by
\begin{equation}
\label{linearcase}
\Phi(x;\theta) = W_K \dots W_1 x~,~\theta = (W_1, \dots, W_K)~.
\end{equation}
and the ridge regression $\mathcal{R}(\theta) =\| \theta \|^2$. This model defines a non-convex (and non-concave) loss $\Fem(\theta)$.
When $\kappa = 0$, it has been shown in \cite{saxe2013exact} and \cite{kawaguchi2016deep} that in this case, 
every local minima is a global minima.  
We provide here an alternative proof of that result that uses
a somewhat simpler argument and allows for $\kappa > 0$ in the case $K=2$.


\begin{proposition}
\label{proplinear}
Let $W_1, W_2, \dots, W_K$ be weight matrices of sizes 
$n_k \times n_{k+1}$, $k < K$, and let $\Fem(\theta)$, $\Forr(\theta)$ 
denote the risk minimizations using $\Phi$ as in (\ref{linearcase}). 
Assume that $n_j > \min(n_1, n_K)$ for $j=2 \dots K-1$.
Then $\Omega_{\Fem}(\lambda)$ (and $\Omega_{\Forr}$) is connected for all $\lambda$ and all $K$ when $\kappa=0$, and for $\kappa>0$ when $K=2$; and therefore there are no poor local minima in these cases. 
Moreover, any $\theta$ can be connected to the lowest energy level with a strictly decreasing path. 
\end{proposition}

Let us highlight that this result is slightly complementary than that of \cite{kawaguchi2016deep}, Theorem 2.3.
Whereas we require $n_j > \min(n_1, n_K)$ for $j=2 \dots K-1$ and our analysis does not inform about the order of the saddle points, 
we do not need full rank assumptions on $\Sigma_X$ nor the weights $W_k$. 

This result does also highlight a certain mismatch between the picture of having no poor local minima 
and generalization error. Incorporating regularization drastically changes the topology, and the 
fact that we are able to show connectedness only in the two-layer case with ridge regression is profound; we conjecture that extending it to deeper models requires a different regularization, perhaps using more general atomic norms \cite{bach2013convex}. But we now move our interest to the nonlinear case, which is more relevant to our purposes. 

\subsection{Half-Rectified Nonlinear Case}

We now study the setting given by 
\begin{equation}
\label{relucase}
\Phi(x;\theta) = W_K \rho W_{K-1} \rho \dots \rho W_1 x~,~\theta = (W_1, \dots, W_K)~,
\end{equation}
where $\rho(z) = \max(0 ,z)$. 
The biases can be implemented by replacing the input vector $x$ 
with $\overline{x}=(x, 1)$ and by rebranding each parameter matrix as 
$$\overline{W}_i = \left( 
\begin{array}{c|c}
W_i & b_i \\
\hline 
0 & 1 
\end{array}
\right)~,$$
where $b_i$ contains the biases for each layer.	
For simplicity, we continue to use $W_i$ and $x$ in the following.

\subsubsection{Nonlinear models are generally disconnected}
\label{disconnect}

One may wonder whether the same phenomena of global connectedness also holds 
in the half-rectified case. A simple motivating counterexample shows that this is not the case in 
general. Consider a simple setup with $X \in \R^2$ drawn from a mixture of two Gaussians $\mathcal{N}_{-1}$ 
and $\mathcal{N}_{1}$, and let $Y = (X-\mu_Z) \cdot Z $ , where $Z$ is the (hidden) mixture component taking $\{1,-1\}$ values.  Let 
$\hat{Y} = \Phi(X; \{ W_1, W_2\} )$ be a single-hidden layer ReLU network, with two hidden units. 
Let $\theta^A$ be a configuration that bisects the two mixture components, 
and let $\theta^B$ the same configuration, but swapping the bisectrices. 
One can verify that they can both achieve arbitrarily small risk by letting the covariance of the mixture components go to $0$. 
However, any path that connects $\theta^A$ to $\theta^B$ 
must necessarily pass through a point in which $W_1$ has rank $1$, which leads to an estimator with risk at least $1/2$.  

In fact, it is easy to see that this counter-example can be extended to any generic half-rectified architecture, if one is 
allowed to adversarially design a data distribution. For any given $\Phi(X; \theta)$ with arbitrary architecture and current parameters 
$\theta = (W_i)$, let $\mathcal{P}_\theta=\{ \mathcal{A}_1, \dots, \mathcal{A}_S\}$ be the underlying tessellation of the input space given by our current choice of parameters; that is, $\Phi(X; \theta)$ is piece-wise linear and $\mathcal{P}_\theta$ contains those pieces. Now let 
$X$ be any arbitrary distribution with density $p(x) > 0$ for all $x \in \R^n$, for example a Gaussian, and let 
$Y ~|~X ~\stackrel{d}{=} \Phi(X ; \theta)$~. Since $\Phi$ is invariant under a subgroup of permutations $\theta_\sigma$ of its hidden layers, it is easy to see that one can find two parameter values $\theta_A = \theta$ and $\theta_B = \theta_\sigma$ such that $\Forr(\theta_A) = \Forr(\theta_B) = 0$, but any continuous path $\gamma(t)$ from $\theta_A$ to $\theta_B$ will have a different tessellation and therefore won't satisfy $\Forr( \gamma(t) ) = 0$. 
Moreover, one can build on this counter-example to show that not only the level sets are disconnected, but also that there exist poor local minima. Let $\theta'$ be a different set of parameters, and $Y' ~|~X \stackrel{d}{=} \Phi(X; \theta')$ be a different target distribution. Now consider the data distribution given by the mixture
$$X ~|~p(x) ~~,~z \sim \text{Bernoulli}(\pi)~,~Y ~|~X,z \stackrel{d}{=} z \Phi(X;\theta) + (1-z) \Phi(X; \theta')~.$$
By adjusting the mixture component $\pi$ we can clearly change the risk at $\theta$ and $\theta'$ and make them different, but we conjecture that this preserves the status of local minima of $\theta$ and $\theta'$. Appendix \ref{sec:disconnect} constructs a counter-example numerically.
 
This illustrates an intrinsic difficulty in the optimization landscape if one is after \emph{universal} 
guarantees that do not depend upon the data distribution. This difficulty is non-existent in the linear case 
and not easy to exploit in mean-field approaches such as \cite{choromanska2015loss}, 
and shows that in general 
we should not expect to obtain connected level sets. However, 
connectedness can be recovered if one is willing to accept a small increase 
of energy and make some assumptions on the complexity of the regression task.
 Our main result shows that the amount by which the energy is 
allowed to increase is upper bounded by a quantity that trades-off model overparametrization 
and smoothness in the data distribution.

For that purpose, we start with a characterization of the oracle loss, and for simplicity let us assume 
$Y \in \R$ and let us first consider the case with a single hidden layer and $\ell_1$ regularization:
$\mathcal{R}(\theta) = \| \theta\|_1$.

\subsubsection{Preliminaries}
 Before proving our main result, we need to introduce  preliminary notation and results. 
We first describe the case with a single hidden layer of size $m$. 

We define
\begin{equation}
\label{bla2}
e(m) = \min_{W_1 \in \R^{m \times n}, \|W_1(i) \|_2 \leq 1, W_2 \in \R^m} \E\{ | \Phi(X; \theta) - Y|^2 \} + \kappa  \| W_2 \|_1~.
\end{equation}
to be the oracle risk using $m$ hidden units with norm $\leq 1$ and using sparse regression. 
It is a well known result by Hornik and Cybenko that a single hidden layer 
is a universal approximator under very mild assumptions, i.e. $\lim_{m \to \infty} e(m) = 0$.
This result merely states that our statistical setup is consistent, and it should not be 
surprising to the reader familiar with classic approximation theory.
 A more interesting question is the rate at which $e(m)$ decays, which depends 
on the smoothness of the joint density $(X, Y) \sim P$ relative to the nonlinear activation 
family we have chosen.

For convenience, we redefine $W = W_1$ and $\beta = W_2$ and
 $Z(W) = \max(0, W X)$. We also write $z(w) = \max(0, \langle w, X \rangle)$ where $(X, Y) \sim P$ and $w \in \R^N$ is any deterministic vector.
Let $\Sigma_X = \E_{P} XX^T \in \R^{N \times N}$ be the covariance operator of the random input $X$. We assume $\| \Sigma_X \| < \infty$. 

 A fundamental property that will be essential to our analysis is that, despite 
the fact that $Z$ is nonlinear, the quantity $[ w_1, w_2 ]_Z := \E_P \{ z(w_1) z(w_2) \} $ 
is locally equivalent to the linear metric $\langle w_1, w_2 \rangle_X = \E_P \{ w_1^T X X^T w_2 \} = \langle w_1, \Sigma_X w_2 \rangle$, and that the linearization error decreases with the angle between $w_1$ and $w_2$. Without loss of generality, we assume here that $\|w_1 \| = \| w_2 \| = 1$, and we write $\| w \|_Z^2 = \E \{ | z(w) |^2 \} $.
\begin{proposition}
\label{localdistprop}
Let $\alpha = \cos^{-1}( \langle w_1, w_2 \rangle )$ be the angle between unitary vectors $w_1$ and $w_2$ and let $w_m =  \frac{w_1 + w_2}{\| w_1 + w_2 \|}$ be their unitary bisector. 
Then
\begin{equation}
\label{localdisteq}
 \frac{1 + \cos \alpha}{2}  \| w_m  \|_Z^2 - 2 \| \Sigma_X \| \left( \frac{1-\cos \alpha}{2} + \sin^2 \alpha \right) \leq [ w_1, w_2 ]_Z \leq \frac{1+\cos \alpha}{2}  \| w_m  \|_Z^2 ~.
\end{equation}
\end{proposition} 
The term $\| \Sigma_X \| $ is overly pessimistic: we can replace it by the energy of $X$ projected into the subspace spanned by $w_1$ and $w_2$ (which is bounded by $2 \| \Sigma_X \|$). 
When $\alpha$ is small, a Taylor expansion of the trigonometric terms reveals that 
\begin{eqnarray*}
\frac{2}{3 \| \Sigma_X \|} \langle w_1, w_2 \rangle &=& \frac{2}{3 \| \Sigma_X \|} \cos \alpha = \frac{2}{3\| \Sigma_X \|}(1 - \frac{\alpha^2}{2} + O(\alpha^4)) \\ 
&\leq& ( 1 - \alpha^2/4)\| w_m \|_Z^2 - \| \Sigma_X \|( \alpha^2/4 + \alpha^2) + O(\alpha^4) \\
&\leq & [ w_1, w_2 ]_Z + O(\alpha^4) ~,
\end{eqnarray*}
and similarly 
$$[ w_1, w_2 ]_Z \leq \langle w_1, w_2 \rangle \| w_m \|_Z^2 \leq \| \Sigma_X\| \langle w_1, w_2 \rangle~.$$
The local behavior of parameters $w_1, w_2$ on our regression problem is thus equivalent to that of having a linear layer, provided $w_1$ and $w_2$ are sufficiently close to each other.
This result can be seen as a \emph{spoiler} of what is coming: increasing the hidden layer dimensionality $m$ will increase the chances to encounter pairs of vectors $w_1, w_2$ with small angle; and with it some hope of approximating the previous linear behavior thanks to the small linearization error. 

In order to control the connectedness, we need a last definition. Given a 
hidden layer of size $m$ with current parameters $W \in \R^{n \times m}$, we define a
``robust compressibility" factor as 
\begin{equation}
\label{compress}
\delta_W(l, \alpha; m) = \min_{ \|\gamma \|_0 \leq l, \sup_i |\angle(\tilde{w}_i, w_i)| \leq \alpha} \E \{| Y - \gamma Z(\tilde{W}) |^2 + \kappa \| \gamma \|_1  \}~,~(l \leq m)~.
\end{equation} 
This quantity thus measures how easily one can compress the current hidden layer representation, 
by keeping only a subset of $l$ its units, but allowing these units to move by a small amount controlled by $\alpha$. It is a form 
of $n$-width similar to Kolmogorov width \cite{donoho2006compressed} and is also related to robust sparse coding from \cite{tang2013compressed, ekanadham2011recovery}.

\subsubsection{Main result}

Our main result considers now a non-asymptotic scenario given by some fixed
size $m$ of the hidden layer. Given two parameter values $\theta^A = (W_1^A, W_2^A) \in \mathcal{W}$ 
and $\theta^B= (W_1^B, W_2^B)$ with $\Forr(\theta^{\{A,B\} } ) \leq \lambda$, 
we show that there exists a continuous path 
$\gamma: [0,1] \to \mathcal{W}$ connecting $\theta^A$ and $\theta^B$ 
such that its oracle risk is uniformly bounded by $\max(\lambda, \epsilon)$, where $\epsilon$ 
decreases with model overparametrization. 
\begin{theorem}
\label{maintheo}
For any $\theta^A, \theta^B \in \mathcal{W}$ and $\lambda \in \R$ satisfying $\Forr(\theta^{\{A,B\}}) \leq \lambda$, there exists a continuous path $\gamma: [0,1] \to \mathcal{W}$ such that
$\gamma(0) = \theta^A$, $\gamma(1) = \theta^B$ and
\begin{equation}
\Forr( \gamma(t) )  \leq \max( \lambda, \epsilon)~,\text{ with}
\end{equation}

\begin{align}
\epsilon = \inf_{l, \alpha} \Bigl(\max \Bigl\{ e(l), &\delta_{W_1^A}(m, 0; m ) , \delta_{W_1^A}(m-l, \alpha; m ) ,   \\ 
&\delta_{W_1^B}(m, 0; m ) ,\delta_{W_1^B} (m-l, \alpha; m ) \Bigr\} + C_1 \alpha  + O(\alpha^2) \Bigr)~,
\end{align}
where $C_1$ is an absolute constant depending only on $\kappa$ and $P$.
\end{theorem}
Some remarks are in order. First, our regularization term is currently a mix between $\ell_2$ norm constraints on the first layer and $\ell_1$ norm constraints on the second layer. We believe this is an artifact of our proof technique, and we conjecture that more general regularizations yield similar results. Next, this result uses the data distribution through the oracle bound $e(m)$ and the covariance term. The 
extension to empirical risk is accomplished by replacing the probability measure $P$ by the empirical measure $\hat{P} = \frac{1}{L} \sum_l \delta\left( (x,y) - (x_l, y_l)\right) $. However, our asymptotic analysis has to be carefully reexamined to take into account and avoid the trivial regime when $M$ outgrows $L$.  
A consequence of Theorem \ref{maintheo} is that as $m$ increases, the model becomes asymptotically connected, as proven in the following corollary.
\begin{corollary}
\label{maincoro}
As $m$ increases, the energy gap $\epsilon$ satisfies $\epsilon = O( m^{-\frac{1}{n}})$ and therefore the level sets become connected at all energy levels.
\end{corollary}
This is consistent with the overparametrization results from \cite{safran2015quality,shamir2} and the general common knowledge amongst deep learning practitioners. Our next sections explore this question, and refine it by considering not only topological properties but also some rough geometrical measure of the level sets.

%% file: geometry.tex
\section{Geometry of Level Sets}
\label{sec:QuanNoncon}

\subsection{The Greedy Algorithm}
\label{sec:GreedyAlg}
 
 The intuition behind our main result is that, for smooth enough loss functions and for sufficient overparameterization, it should be ``easy'' to connect two equally powerful models---i.e., two models with $F_o{\theta^{A,B}} \leq \lambda$.  A sensible measure of this ease-of-connectedness is the normalized length of the geodesic connecting one model to the other: $|\gamma_{A,B}(t)| / |\theta_A - \theta_B|$.  This length represents approximately how far of an excursion one must make in the space of models relative to the euclidean distance between a pair of models.  Thus, convex models have a geodesic length of $1$, because the geodesic is simply linear interpolation between models, while more non-convex models have geodesic lengths strictly larger than $1$.
 
 Because calculating the exact geodesic is difficult, we approximate the geodesic paths via a dynamic programming approach we call Dynamic String Sampling.  We comment on alternative algorithms in Appendix \ref{sec:ConstrainedAlg}.
 
 For a pair of models with network parameters $\theta_i$, $\theta_j$, each with $F_e(\theta)$ below a threshold $L_0$, we aim to efficienly generate paths in the space of weights where the empirical loss along the path remains below $L_0$.  These paths are continuous curves belonging to $\Omega_F(\lambda)$--that is, the level sets of the loss function of interest.

\begin{algorithm}
\caption{Greedy Dynamic String Sampling}\label{euclid}
\begin{algorithmic}[1]
{\scriptsize 
\State $\text{$L_0$} \gets \text{Threshold below which path will be found}$
\State $\text{$\Phi_1$} \gets \text{randomly initialize } $$\theta_1$$ \text{, train } $$\Phi (x_i\;\theta_1)$$ \text{ to $L_0$}$
\State $\text{$\Phi_2$} \gets \text{randomly initialize } $$\theta_2$$ \text{, train } $$\Phi (x_i\;\theta_2)$$ \text{ to $L_0$}$

\State $\text{BeadList} \gets $$(\Phi_1,\Phi_2)$
\State $\text{Depth} \gets 0$ 

\Procedure{FindConnection}{$\Phi_1,\Phi_2$}
\State $\text{$t^*$} \gets \text{t such that } $$\frac{d \gamma(\theta_1, \theta_2, t)}{dt} \bigg|_{t} = 0$$  \text{ OR } $$t = 0.5$$ $
\State $\text{$\Phi_3$} \gets \text{train } $$\Phi(x_i; t^*\theta_1 + (1-t^*)\theta_2)$$ \text{ to $L_0$}$
\State $\text{BeadList} \gets \text{insert}$$(\Phi_3$$\text{, after } $$\Phi_1$$\text{, BeadList)}$
\State $\text{$MaxError_1$} \gets \text{$max_t$}$$(F_e(t\theta_3 + (1-t)\theta_1))$$ $
\State $\text{$MaxError_2$} \gets \text{$max_t$}$$(F_e(t\theta_2 + (1-t)\theta_3))$$ $
\If {$\text{$MaxError_1$} > \text{$L_0$ }} \text{ }\Return \text{ FindConnection}$$(\Phi_1,\Phi_3)$$ $
\EndIf
\If {$\text{$MaxError_2$} > \text{$L_0$ }} \text{ }\Return \text{ FindConnection}$$(\Phi_3,\Phi_2)$$ $
\EndIf
\State $\text{Depth} \gets \text{Depth$+1$}$ 
\EndProcedure }
\end{algorithmic}
\end{algorithm}

  The algorithm recursively builds a string of models in the space of weights which continuously connect $\theta_i$ to $\theta_j$.  Models are added and trained until the pairwise linearly interpolated loss, i.e. $\rm{max}_t F_e(t\theta_i\ +\ (1-t)\theta_j)$ for $t\in(0,1)$, is below the threshold, $L_0$, for every pair of neighboring models on the string.  We provide a cartoon of the algorithm in Appendix \ref{AlgCartoon}.

  \subsection{Failure Conditions and Practicalities}
  \label{sec:Fail}
  
  While the algorithm presented will faithfully certify two models are connected if the algorithm converges, it is worth emphasizing that the algorithm does not guarantee that two models are disconnected if the algorithm fails to converge.  In general, the problem of determining if two models are connected can be made arbitrarily difficult by choice of a particularly pathological geometry for the loss function, so we are constrained to heuristic arguments for determining when to stop running the algorithm.  Thankfully, in practice, loss function geometries for problems of interest are not intractably difficult to explore.  We comment more on diagnosing disconnections more carefully in Appendix \ref{sec:disconnect}.
  
  Further, if the $\rm{\mathbf{MaxError}}$ exceeds $L_0$ for every new recursive branch as the algorithm progresses, the worst case runtime scales as $O(\rm{exp}(\rm{\mathbf{Depth}}))$.  Empirically, we find that the number of new models added at each depth does grow, but eventually saturates, and falls for a wide variety of models and architectures, so that the typical runtime is closer to $O(\rm{poly}(\rm{\mathbf{Depth}}))$---at least up until a critical value of $L_0$.
  
  To aid convergence, either of the choices in line $7$ of the algorithm works in practice---choosing $t^*$ at a local maximum can provide a modest increase in algorithm runtime, but can be unstable if the the calculated interpolated loss is particularly flat or noisy.  $t^*=.5$ is more stable, but slower.  Finally, we find that training $\Phi_3$ to $\alpha L_0$ for $\alpha < 1$ in line $8$ of the algorithm tends to aid convergence without noticeably impacting our numerics.  We provide further implementation details in \ref{sec:NumExp}.

%% file: experiments.tex
\section{Numerical Experiments}
\label{sec:NumExp}

For our numerical experiments, we calculated normalized geodesic lengths for a variety of regression and classification tasks.  In practice, this involved training a pair of randomly initialized models to the desired test loss value/accuracy/perplexity, and then attempting to connect that pair of models via the Dynamic String Sampling algorithm.  We also tabulated the average number of ``beads'', or the number intermediate models needed by the algorithm to connect two initial models.  For all of the below experiments, the reported losses and accuracies are on a restricted test set.  For more complete architecture and implementation details, see our \href{github.com/danielfreeman11/convex-nets}{GitHub page}.

The results are broadly organized by increasing model complexity and task difficulty, from easiest to hardest.  Throughout, and remarkably, we were able to easily connect models for every dataset and architecture investigated except the one explicitly constructed counterexample discussed in Appendix \ref{symdisc}.  Qualitatively, all of the models exhibit a transition from a highly convex regime at high loss to a non-convex regime at low loss, as demonstrated by the growth of the normalized length as well as the monotonic increase in the number of required ``beads'' to form a low-loss connection.

%

\subsection{Polynomial Regression}
\label{sec:PolyFuncs}

\begin{figure} 
  \centering
\begin{subfigure}[b]{.34\textwidth}
\includegraphics[width=\textwidth]{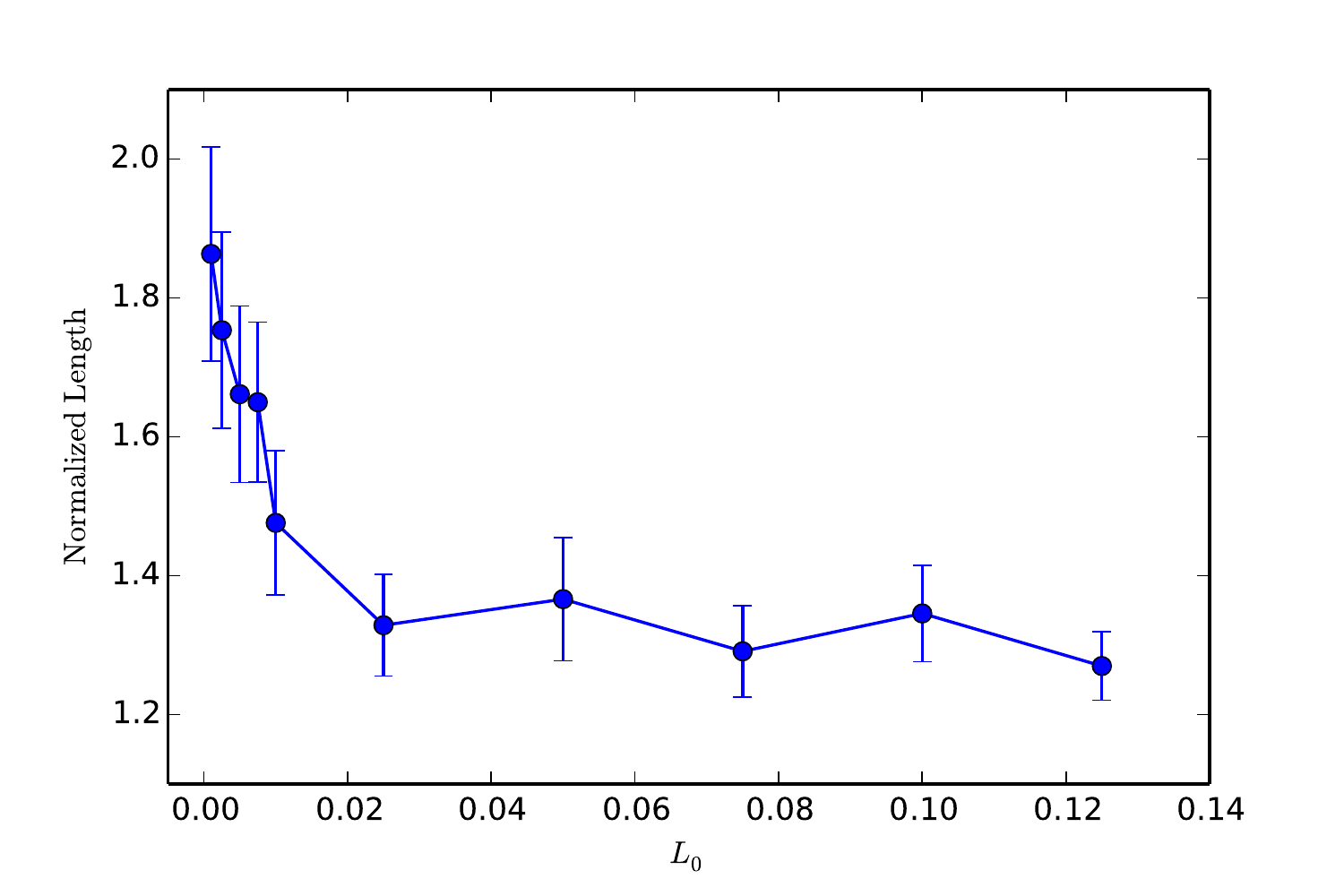}
\caption*{(1a)}
\end{subfigure}
\begin{subfigure}[b]{.34\textwidth}
\includegraphics[width=\textwidth]{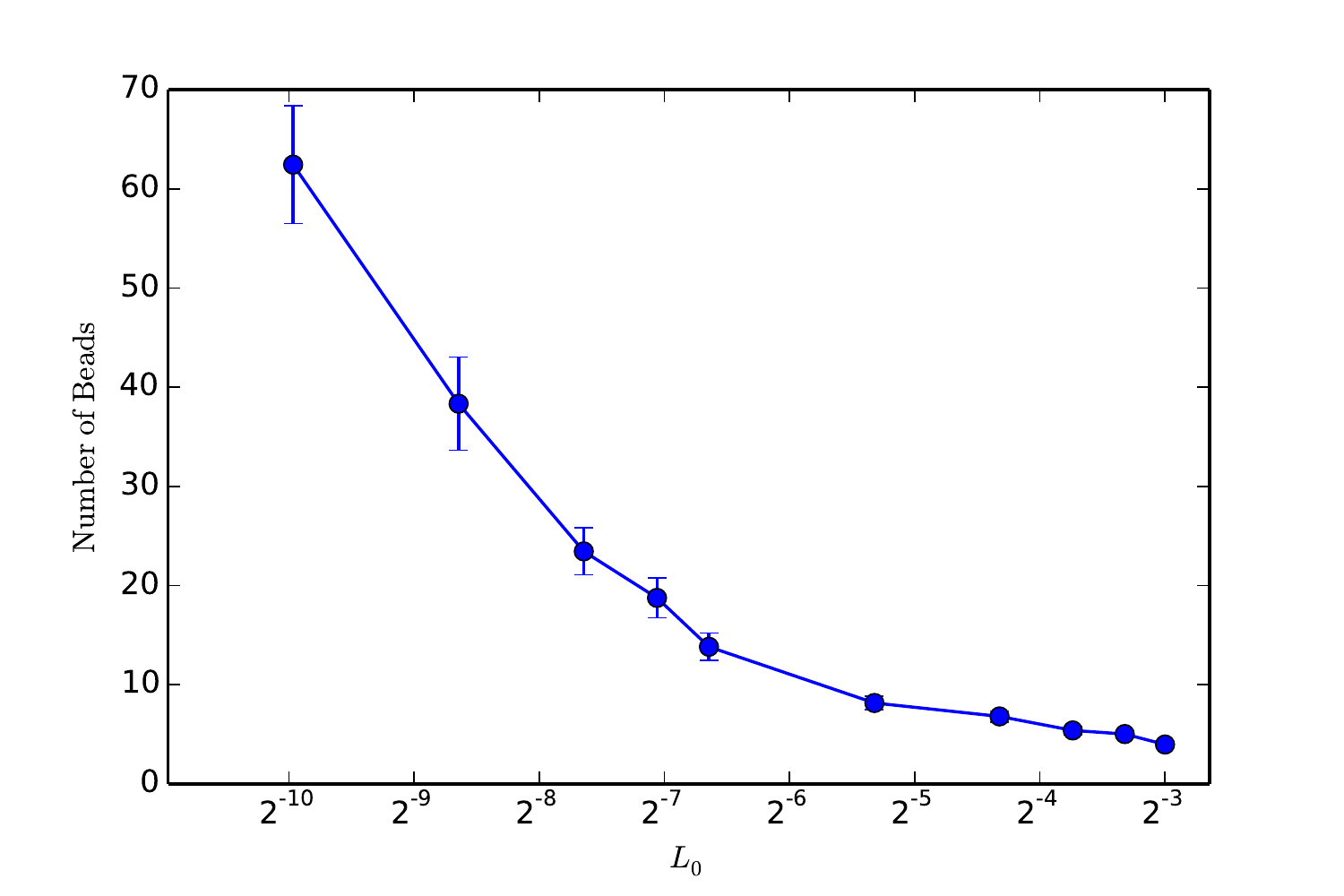}
\caption*{(1b)} 
\end{subfigure}

\begin{subfigure}[b]{.34\textwidth}
\includegraphics[width=\textwidth]{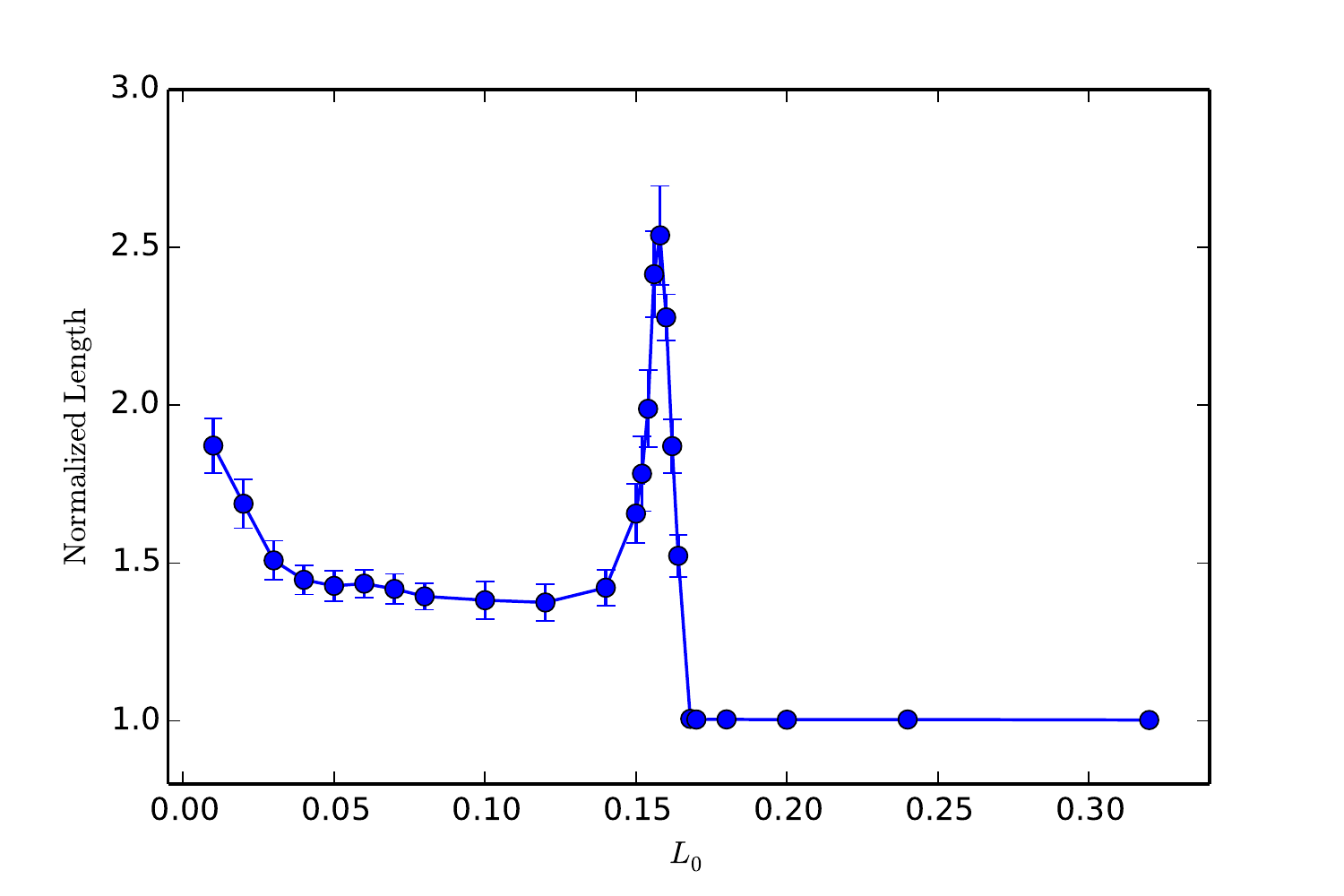} 
\caption*{(2a)}
\end{subfigure}
\begin{subfigure}[b]{.34\textwidth}
\includegraphics[width=\textwidth]{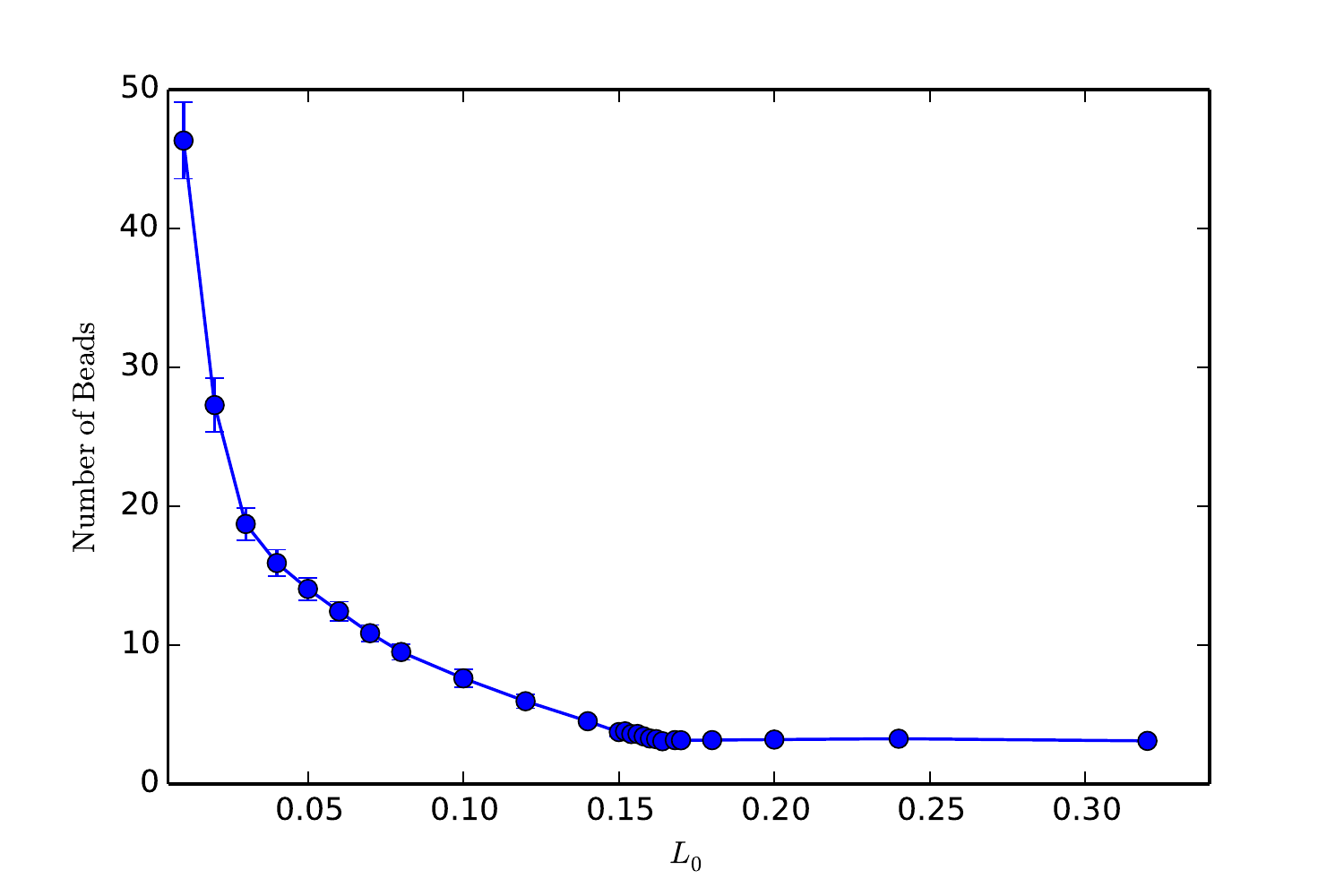}
\caption*{(2b)}
\end{subfigure}

\begin{subfigure}[b]{.34\textwidth}
\includegraphics[width=\textwidth]{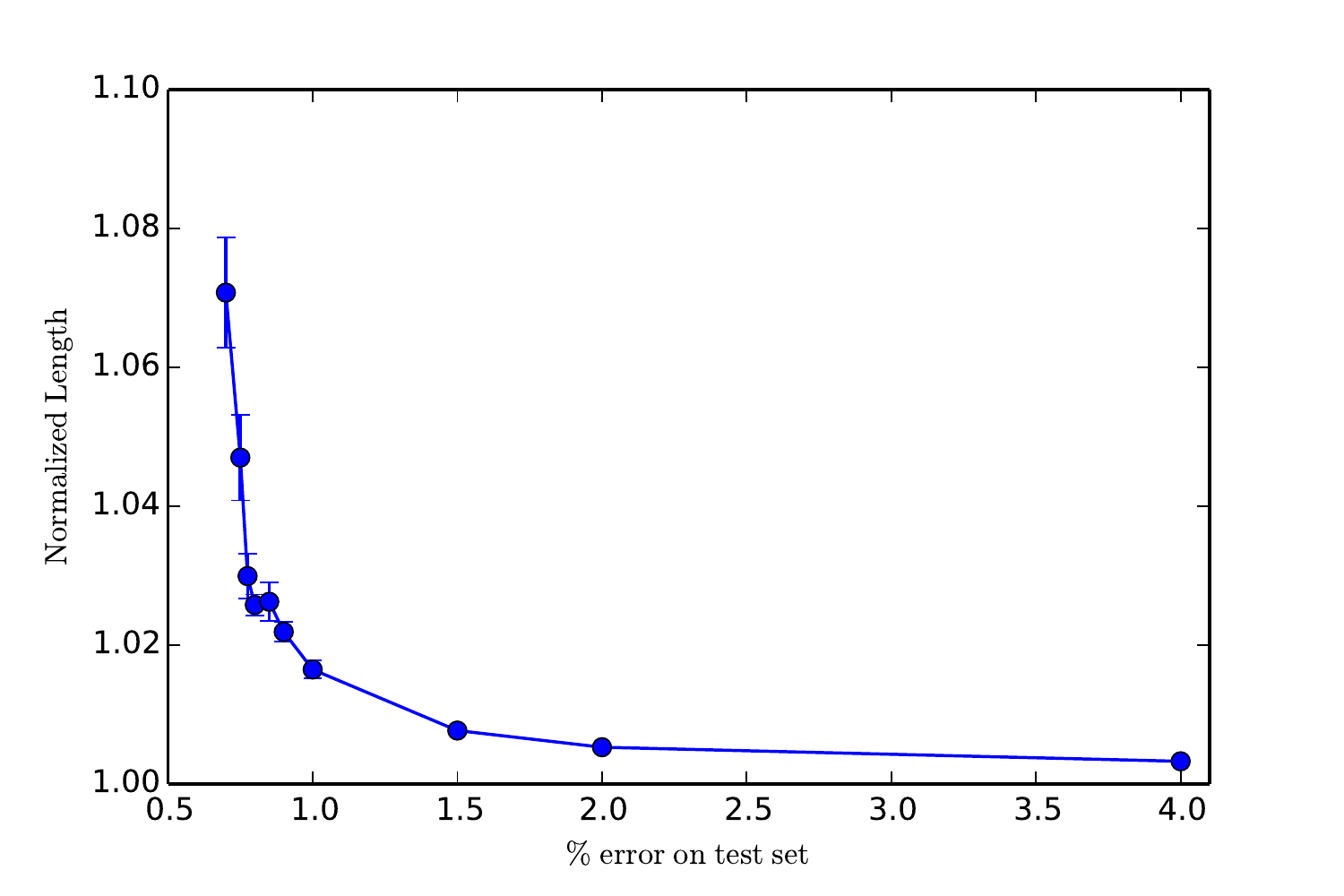}
\caption*{(3a)} 
\end{subfigure}
\begin{subfigure}[b]{.34\textwidth}
\includegraphics[width=\textwidth]{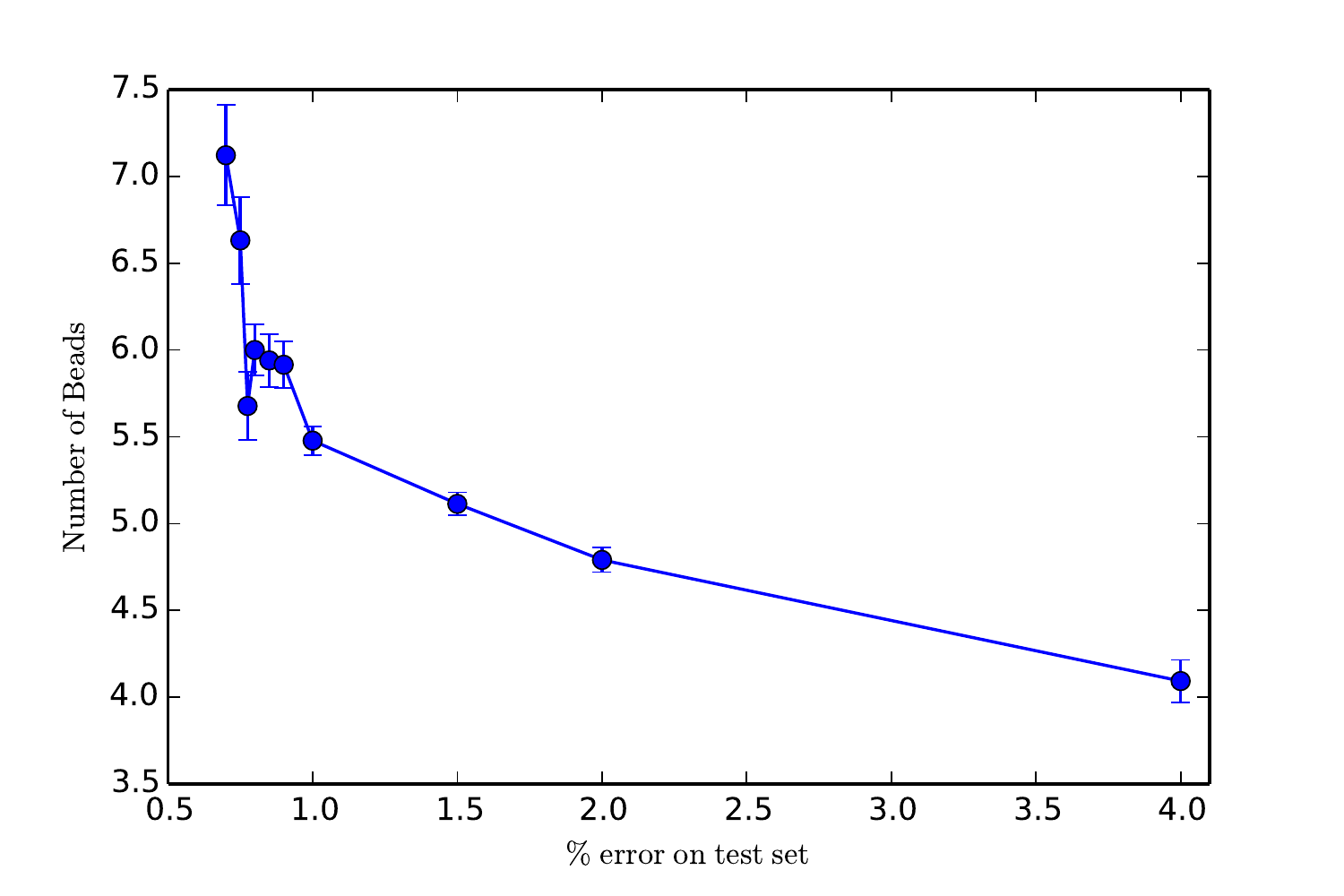}
\caption*{(3b)}
\end{subfigure}

\begin{subfigure}[b]{.34\textwidth}
\includegraphics[width=\textwidth]{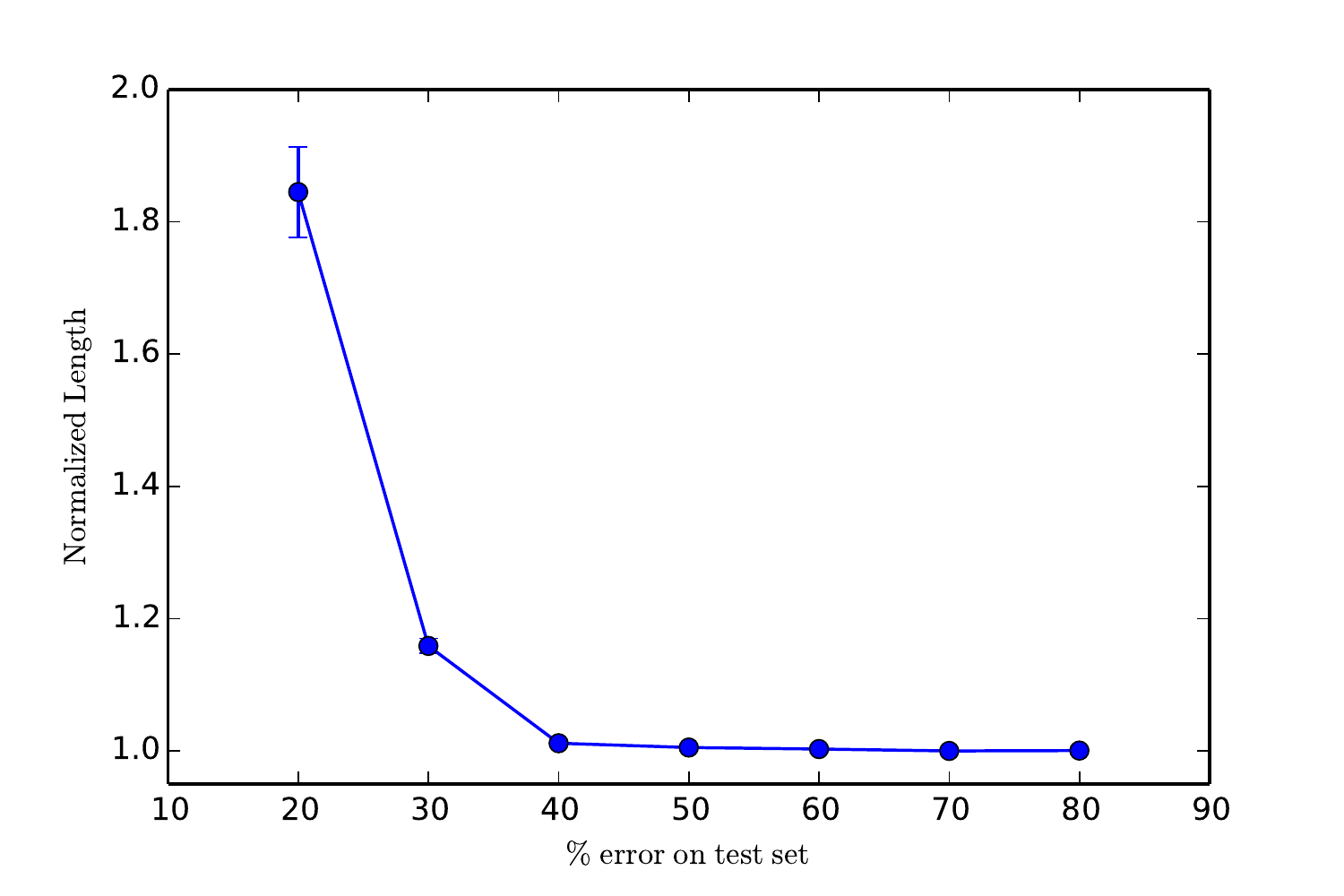}
\caption*{(4a)}
\end{subfigure}
\begin{subfigure}[b]{.34\textwidth}
\includegraphics[width=\textwidth]{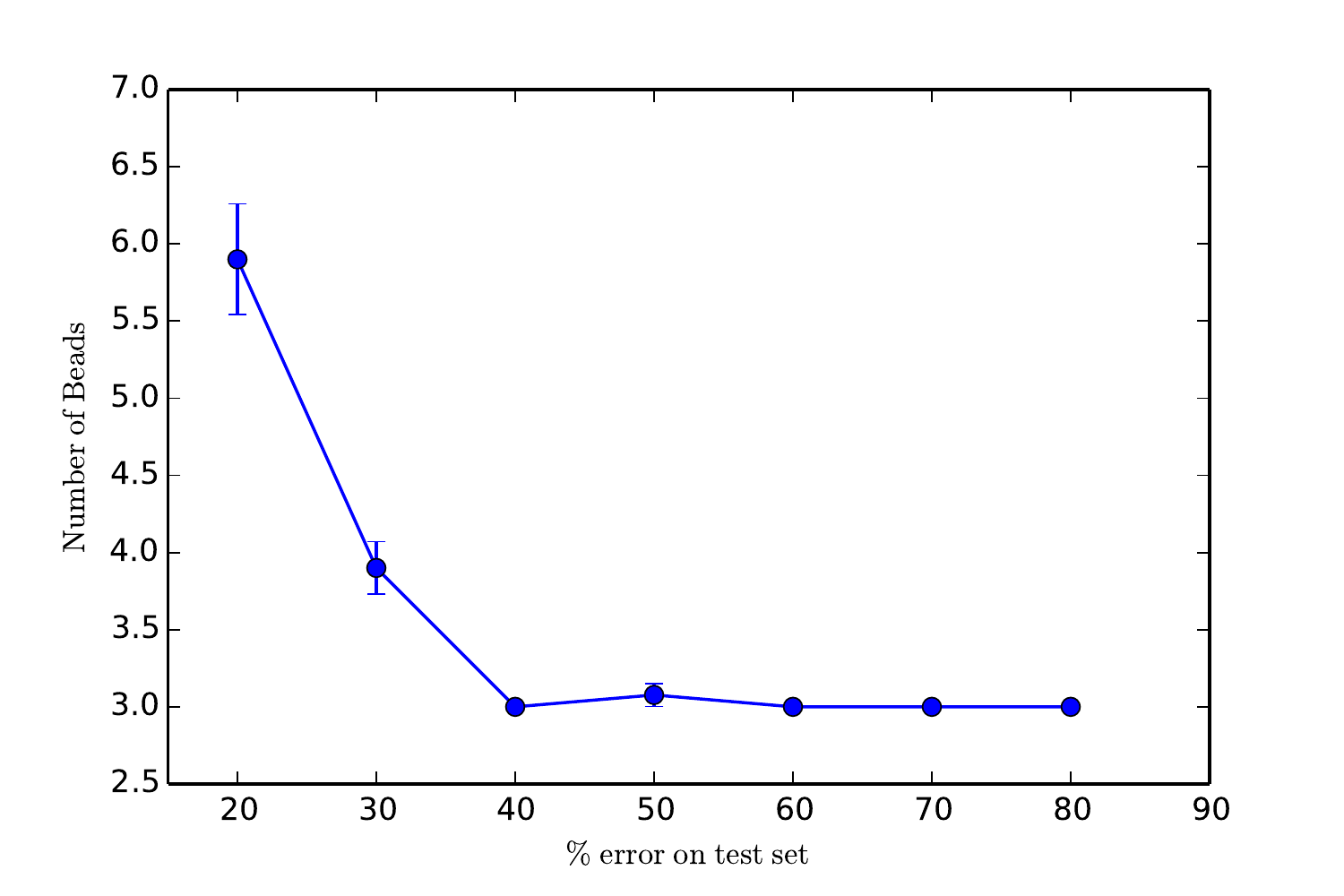}
\caption*{(4b)} 
\end{subfigure}

\begin{subfigure}[b]{.34\textwidth}
\includegraphics[width=\textwidth]{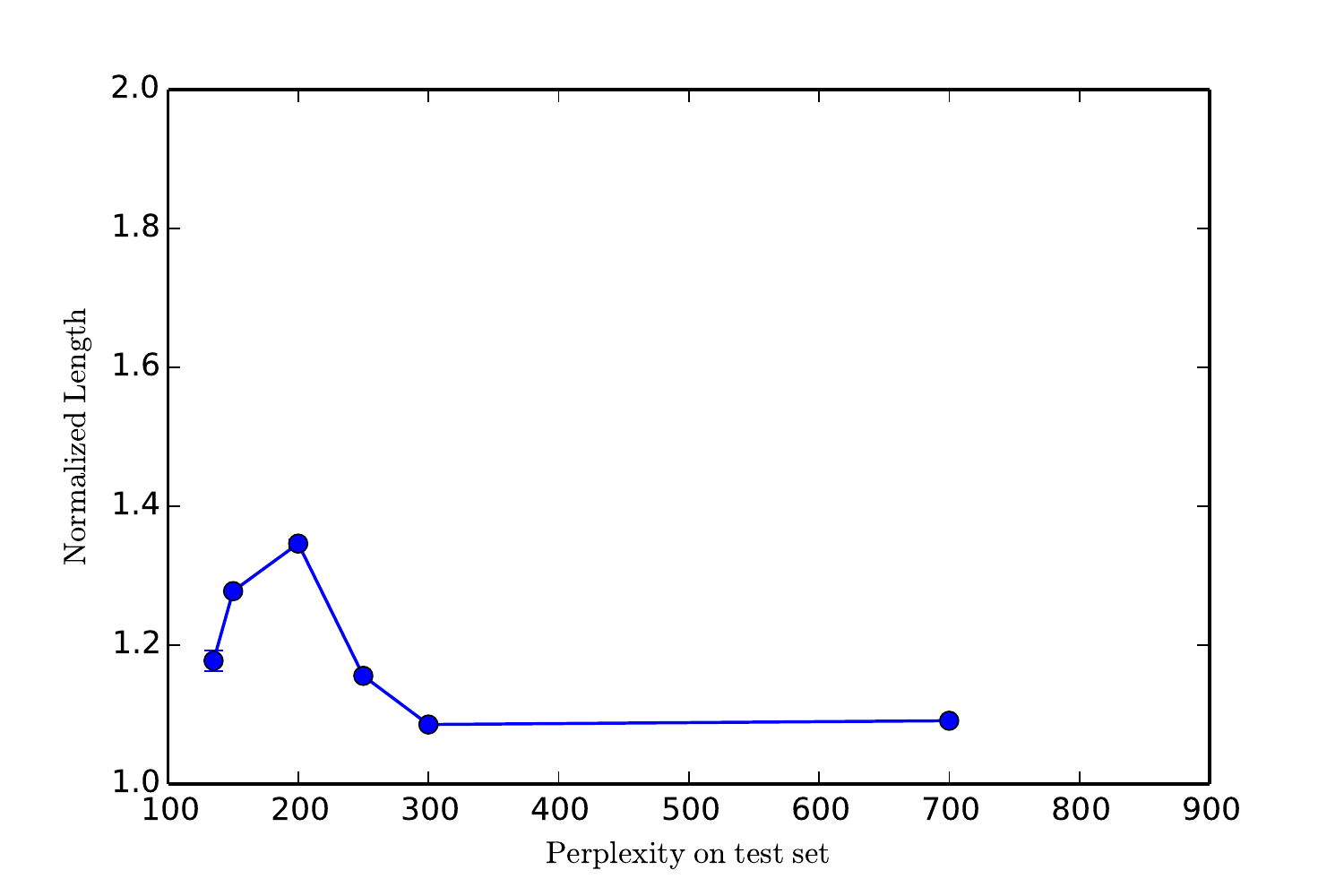}
\caption*{(5a)} 
\end{subfigure}
\begin{subfigure}[b]{.34\textwidth}
\includegraphics[width=\textwidth]{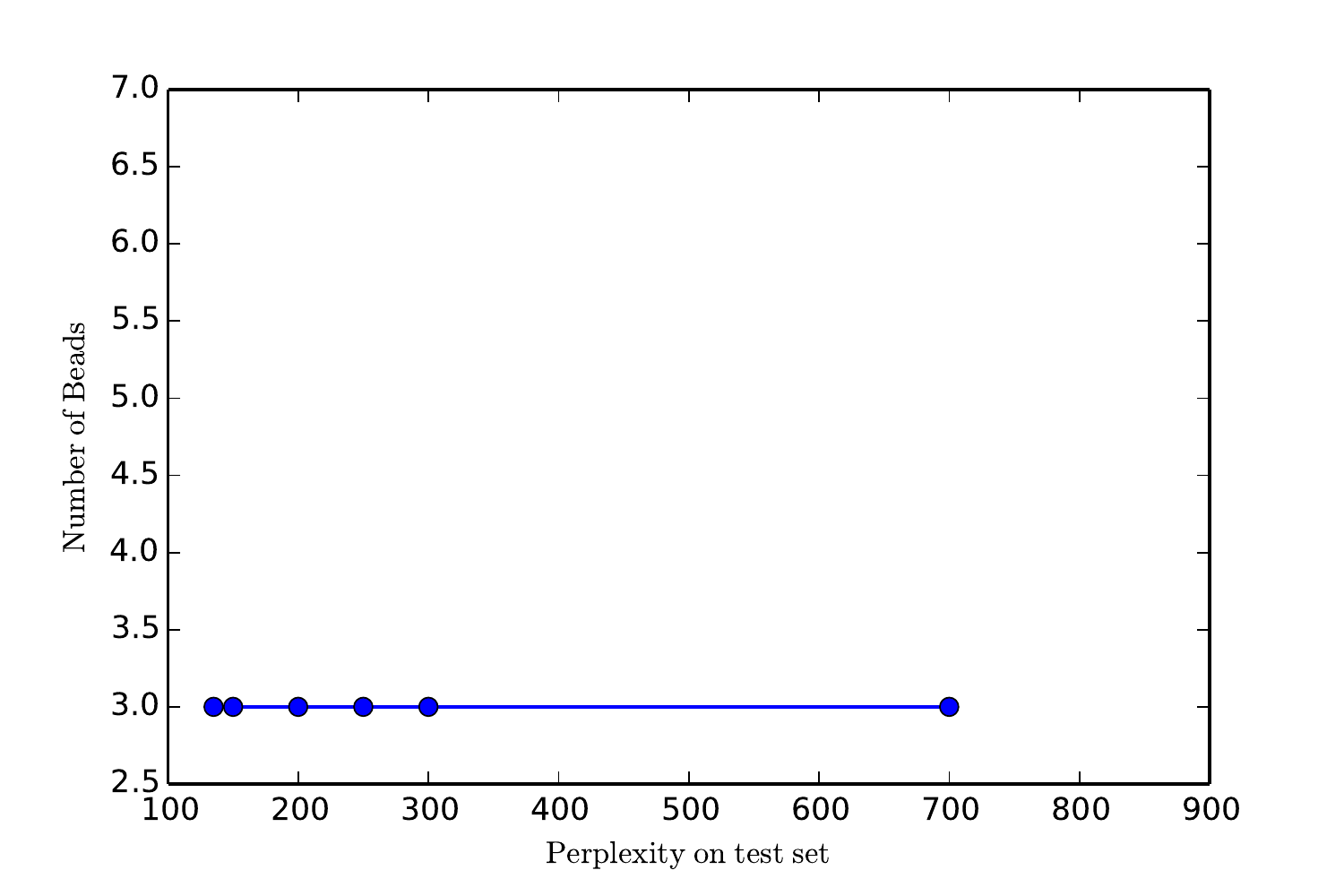}
\caption*{(5b)}
\end{subfigure}
   
  \caption{(Column a) Average normalized geodesic length and (Column b) average number of beads versus loss. (1) \href{github.com/danielfreeman11/convex-nets/tree/master/LaunchScripts/QUADRATIC.py}{A quadratic regression task}. (2) \href{github.com/danielfreeman11/convex-nets/tree/master/LaunchScripts/CUBIC.py}{A cubic regression task}. (3) A convnet for \href{github.com/danielfreeman11/convex-nets/tree/master/LaunchScripts/MNIST.py}{MNIST}. (4) A convnet inspired by \href{www.cs.toronto.edu/\%7Ekriz/cifar.html}{Krizhevsky} for \href{github.com/danielfreeman11/convex-nets/tree/master/LaunchScripts/CIFAR10.py}{CIFAR10}. (5) \href{github.com/danielfreeman11/convex-nets/tree/master/LaunchScripts/PTBRNN.py}{A RNN} inspired by \href{arxiv.org/pdf/1409.2329.pdf}{Zaremba} for PTB next word prediction.}
  \label{FigTable}
\end{figure}

 We studied a 1-4-4-1 fully connected multilayer perceptron style architecture with sigmoid nonlinearities and RMSProp/ADAM optimization.  For ease-of-analysis, we restricted the training and test data to be strictly contained in the interval $x\in[0,1]$ and $f(x)\in[0,1]$.  The number of required beads, and thus the runtime of the algorithm, grew approximately as a power-law, as demonstrated in Table \ref{FigTable} Fig. 1.  We also provide a visualization of a representative connecting path between two models of equivalent power in Appendix \ref{visualization}.

 The cubic regression task exhibits an interesting feature around $L_0=.15$ in Table \ref{FigTable} Fig. 2, where the normalized length spikes, but the number of required beads remains low.  Up until this point, the cubic model is strongly convex, so this first spike seems to indicate the onset of non-convex behavior and a concomitant radical change in the geometry of the loss surface for lower loss.

\subsection{Convolutional Neural Networks}
\label{sec:CNN}

 To test the algorithm on larger architectures, we ran it on the MNIST hand written digit recognition task as well as the CIFAR10 image recognition task, indicated in Table \ref{FigTable}, Figs. 3 and 4.  Again, the data exhibits strong qualitative similarity with the previous models: normalized length remains low until a threshold loss value, after which it grows approximately as a power law.  Interestingly, the MNIST dataset exhibits very low normalized length, even for models nearly at the state of the art in classification power, in agreement with the folk-understanding that MNIST is highly convex and/or ``easy''.  The CIFAR10 dataset, however, exhibits large non-convexity, even at the modest test accuracy of 80\%.

\subsection{Recurrent Neural Networks}

 To gauge the generalizability of our algorithm, we also applied it to an LSTM architecture for solving the next word prediction task on the PTB dataset, depicted in Table \ref{FigTable} Fig. 5.  Noteably, even for a radically different architecture, loss function, and data set, the normalized lengths produced by the DSS algorithm recapitulate the same qualitative features seen in the above datasets---i.e., models can be easily connected at high perplexity, and the normalized length grows at lower and lower perplexity after a threshold value, indicating an onset of increased non-convexity of the loss surface.

%% file: conclusions.tex
\section{Discussion}
\label{sec:Discussion}

We have addressed the problem of characterizing the loss surface of neural networks from the perspective
of gradient descent algorithms. We explored two angles -- topological and geometrical aspects -- that build on top of each other.

On the one hand, we have presented new theoretical results that quantify 
the amount of uphill climbing that is required in order to progress to lower energy configurations in 
single hidden-layer ReLU networks, and proved that this amount converges to zero with overparametrization under mild conditions. On the other hand, we have introduced a dynamic programming algorithm that efficiently approximates geodesics within each level set, providing a tool that not only verifies the connectedness of level sets, but also estimates the geometric regularity of these sets. Thanks to this information, we can quantify how `non-convex' an optimization problem is, and verify that the optimization of quintessential deep learning tasks -- CIFAR-10 and MNIST classification using CNNs, and next word prediction using LSTMs -- behaves in a nearly convex fashion up until they reach high accuracy levels.

That said, there are some limitations to our framework. In particular, we do not address saddle-point issues that can greatly affect the actual convergence of gradient descent methods. There are also a number of open questions; amongst those, in the near future we shall concentrate on:
\begin{itemize}
\item \emph{Extending Theorem \ref{maintheo} to the multilayer case}. We believe this is within reach, since the main analytic tool we use is that small changes in the parameters result in small changes in the covariance structure of the features. That remains the case in the multilayer case. 
\item \emph{Empirical versus Oracle Risk}. A big limitation of our theory is that right now it does not inform us on the differences between optimizing the empirical risk versus the oracle risk. Understanding the impact of generalization error and stochastic gradient in the ability to do small uphill climbs is an open line of research.
\item \emph{Influence of symmetry groups}.  Under appropriate conditions, the presence of discrete symmetry groups does not prevent the loss from being connected, but at the expense of increasing the capacity. An important open question is whether one can improve the asymptotic properties by relaxing connectedness to being connected up to discrete symmetry. 
\item \emph{Improving numerics with Hyperplane method}. Our current numerical experiments employ a greedy (albeit faster) algorithm to discover connected components and estimate geodesics. We plan to perform experiments using the less greedy algorithm described in Appendix \ref{sec:ConstrainedAlg}. 
\end{itemize}

%

%% file: constrained.tex
\section{Constrained Dynamic String Sampling}
  \label{sec:ConstrainedAlg}
  
  While the algorithm presented in Sec. \ref{sec:GreedyAlg} is fast for sufficiently smooth families of loss surfaces with few saddle points, here we present a slightly modified version which, while slower, provides more control over the convergence of the string.  We did not use the algorithm presented in this section for our numerical studies.  
  
  Instead of training intermediate models via full SGD to a desired accuracy as in step $8$ of the algorithm, intermediate models are be subject to a constraint that ensures they are ``close'' to the neighboring models on the string.  Specifically, intermediate models are constrained to the unique hyperplane in weightspace equidistant from its two neighbors.  This can be further modified by additional regularization terms to control the ``springy-ness'' of the string.  These heuristics could be chosen to try to more faithfully sample the geodesic between two models.  
  
  In practice, for a given model on the string, $\theta_i$, these two regularizations augment the standard loss by: $\tilde{F}(\theta) = F(\theta)+\zeta(\|\theta_{i-1} - \theta_i\|+\|\theta_{i+1} - \theta_i\|) + \kappa \|\frac{(\theta_{i-1} - \theta_{i+1})/2}{\|(\theta_{i-1} - \theta_{i+1})/2\|} \cdot \frac{(\theta_i - (\theta_{i-1} - \theta_{i+1})/2)}{\| (\theta_i - (\theta_{i-1} - \theta_{i+1})/2)\|}\|$.  The $\zeta$ regularization term controls the ``springy-ness'' of the weightstring, and the $\kappa$ regularization term controls how far off the hyperplane a new model can deviate.  
  
  Because adapting DSS to use this constraint is straightforward, here we will describe an alternative ``breadth-first'' approach wherein models are trained in parallel until convergence.  This alternative approach has the advantage that it will indicate a disconnection between two models ``sooner'' in training.  The precise geometry of the loss surface will dictate which approach to use in practice.
  
  Given two random models $\sigma_i$ and $\sigma_j$ where $|\sigma_i - \sigma_j| < L_0$, we aim to follow the evolution of the family of models connecting $\sigma_i$ to $\sigma_j$.  Intuitively, almost every continuous path in the space of random models connecting $\sigma_i$ to $\sigma_j$ has, on average, the same (high) loss.  For simplicity, we choose to initialize the string to the linear segment interpolating between these two models.  If this entire segment is evolved via gradient descent, the segment will either evolve into a string which is entirely contained in a basin of the loss surface, or some number of points will become fixed at a higher loss.  These fixed points are difficult to detect directly, but will be indirectly detected by the persistence of a large interpolated loss between two adjacent models on the string.
  
  The algorithm proceeds as follows:
  
  (0.) Initialize model string to have two models, $\sigma_i$ and $\sigma_j$.
  
  1. Begin training all models to the desired loss, keeping the instantaneous loss, $L_0(t)$, of all models being trained approximately constant.
  
  2. If the pairwise interpolated loss between $\sigma_n$ and $\sigma_{n+1}$ exceeds $L_0(t)$, insert a new model at the maximum of the interpolated loss (or halfway) between these two models.
  
  3. Repeat steps (1) and (2) until all models (and interpolated errors) are below a threshold loss $L_0(t_{\rm{final}}):=L_0$, or until a chosen failure condition (see \ref{sec:Fail}).

%% file: proofs.tex
\section{Proofs}

\subsection{Proof of Proposition \ref{connectedminima}}

Suppose that $\theta_1$ is a local minima and $\theta_2$ is a global minima, 
but $F(\theta_1) > F(\theta_2)$. If $\lambda = F(\theta_1)$, then clearly 
$\theta_1$ and $\theta_2$ both belong to $\Omega_F(\lambda)$. Suppose 
now that $\Omega_F(\lambda)$ 
is connected. Then we could find a smooth (i.e. continuous and differentiable) path $\gamma(t)$ 
with $\gamma(0) = \theta_1$, $\gamma(1)= \theta_2$ and $F(\gamma(t)) \leq \lambda = F(\theta_1)$.
But this contradicts the strict local minima status of $\theta_1$, and therefore $\Omega_F(\lambda)$ cannot be connected $\square$.


%

\subsection{Proof of Proposition \ref{proplinear}}

Let us first consider the case with $\kappa =0$.
We proceed by induction over the number of layers $K$. 
For $K=1$, the loss $F(\theta)$ is convex. Let  $\theta^A$, $\theta^B$ be two arbitrary points 
in a level set $\Omega_\lambda$. Thus $F(\theta^A) \leq \lambda$ and $F(\theta^B) \leq \lambda$. By definition
of convexity, a linear path is sufficient in that case to connect $\theta^A$ and $\theta^B$:
$$F( (1-t) \theta^A + t \theta^B) \leq (1-t) F(\theta^A) + t F(\theta^B) \leq \lambda~.$$
Suppose the result is true for $K-1$. Let $\theta^A = (W_1^A, \dots, W^A_K)$ and 
 $\theta^B = (W_1^B, \dots, W^B_K)$ with $F(\theta^A) \leq \lambda$, $F(\theta^B) \leq \lambda$.
 Since $n_j \geq \min(n_1, n_K)$ for $j=2 \dots K-1$, we can find $k^*=\{1, K-1\}$ such that
 $n_{k^*} \geq \min(n_{k^*-1}, n_{k^*+1})$.
For each $W_1, \dots, W_K$, we denote $\tilde{W}_j = W_j$ for $j \neq k^*, k^*- 1$ and
$\tilde{W}_{k^*} = W_{k^*-1} W_{k^*}$. 
By induction hypothesis, the 
loss expressed in terms of $\tilde{\theta} = (\tilde{W}_1, \dots, \tilde{W}_{K-1})$ is connected 
between $\tilde{\theta}^A$ and $\tilde{\theta}^B$. Let $\tilde{W}_{k^*}(t)$ the corresponding 
linear path projected in the  layer $k^*$. 
We need to produce a path in the variables $W_{k^*-1}(t)$, $W_{k^*}(t)$ 
such that:
\begin{itemize}
\item[i] $W_{k^*-1}(0) = W_{k^*-1}^A$, $W_{k^*-1}(1) = W_{k^*-1}^B$, 
\item[ii] $W_{k^*}(0) = W_{k^*}^A$, $W_{k^*}(1) = W_{k^*}^B$,
\item[iii] $W_{k^*}(t) W_{k^*-1}(t) = \tilde{W}_{k^*-1}(t) $ for $t \in (0,1)$. 
\end{itemize} 

For simplicity, we denote by $n$ and $m$ the dimensions of $\tilde{W}_{k^*}(t)$, and 
assume without loss of generality that $n \geq m$. 

Suppose first that $\text{rank}(W_{k^*-1}^A) = \text{rank}(W_{k^*-1}^B) =m$. Hence $\min( \lambda_{min}(W_{k^*-1}^A),\lambda_{min}(W_{k^*-1}^B)) = \rho > 0 $.
Let $W_{k^*-1}^A = U_A S_A V_A^T$,  $W_{k^*-1}^B = U_B S_B V_B^T$ be the singular value decomposition of 
$W_{k^*-1}^A$ and $W_{k^*-1}^B$ respectively, with $V_{\{A,B\}} \in \R^{m \times m}$. Observe that by appropriately 
flipping the signs of columns of $V$ and $U$, we can always assume that $\text{det}(V_A) = \text{det}(V_B) =1$. 
Since $GL(\R^m)$ has two connected components and $V_A$ and $V_B$ belong to the same one, we can find 
a continuous path $t\in [0,1] \mapsto V(t) \in GL(\R^m)$ with $V(0) = V_A$, $V(1) = V_B$ and $\text{det}(V(t)) = 1$ for all $t$. 
Also, since $n_1 > m$ by assumption, we can always complete the rectangular matrices $U_{\{A,B\}} \in \R^{n_1 \times m}$ into
$\bar{U}_{\{A,B\}} \in \R^{n_1 \times n_1}$, such that $\text{det}(\bar{U}_A) = \text{det}(\bar{U}_B) = 1$. It follows that we can also 
consider a path $t \mapsto \bar{U}(t)$ with $\bar{U}(0) = \bar{U}_A$, $\bar{U}(1) = \bar{U}_B$ and $\text{det}(\bar{U}(t)) = 1$ for all $t$. 
In particular, since $\text{rank}(\bar{U}(t)) = n_1$ for all $t\in[0,1]$, the restriction of $\bar{U}(t)$ to its first $m$ columns, $U(t)$, has rank $m$ for all $t$. 
Finally, since the singular values $s_{A,1} \dots, s_{A,m}$, $s_{B,1} \dots, s_{B,m}$ are lower bounded by $\rho > 0$, 
we can construct a path $t \mapsto S(t)$ such that $S(t)$ is diagonal, $S(0) = S_A$, $S(1) = S_B$, and $S(t)_{i,i} \geq \rho >0$ for all $t \in [0,1]$.

We consider the path 
\begin{equation}
t \mapsto~W_{k^*-1}(t) = U(t) S(t) V(t)^T~.
\end{equation}


$W_{k^*-1}(t)$ has the property that $W_{k^*-1}(0) = W_{k^*-1}^A$, $W_{k^*-1}(1) = W_{k^*-1}^B$.  
Thanks to the fact that $\text{rank}(W_{k^*-1}(t)) = m$ for all $t \in (0,1)$, there exists $W_{k^*}(t)$ such that
\begin{equation}
\label{lem}
\forall t \in (0,1)~,~ \tilde{W}_{k^*}(t) = W_{k^*}(t) W_{k^*-1}(t)~.
\end{equation}
%

Finally, we need to show that the path $W_{k^*}(t)$ is continuous 
and satisfies $W_{k^*}(0) = W_{k^*}^A$, $W_{k^*}(1) = W_{k^*}^B$.
Since by construction the paths are continuous in $t \in (0,1)$, it only remains to be shown that
\begin{equation}
\lim_{t\to 0} W_{k^*}(t) = W_{k^*}^A~,~\lim_{t\to 1} W_{k^*}(t) = W_{k^*}^B~.
\end{equation}

From (\ref{lem}) we have 
$$W_{k^*}(t) = \tilde{W}_{k^*}(t) W_{k^*-1}(t)^{-1} ~. $$
Consider first the case $t \to 0$.
Since $\tilde{W}_{k^*}(t)$ is continuous in a compact interval,  
we have $\sup_t \|\tilde{W}_{k^*}(t) \| < \infty$. Also, $\|W_{k^*-1}(t)^{-1}\| = \| S_A(t) \|^{-1} < \rho^{-1} $, 
so we have 
\begin{eqnarray}
\label{kem}
&& \lim_{t \to 0} \|  \tilde{W}_{k^*}(t)  W_{k^*-1}(t)^{-1}  - W_{k^*}^A  \| =  \nonumber \\
&=& \lim_{t \to 0} \|  \tilde{W}_{k^*}(t)  W_{k^*-1}(t)^{-1}  -  \tilde{W}_{k^*}(t)  (W_{k^*-1}^A)^{-1} +  \tilde{W}_{k^*}(t)  (W_{k^*-1}^A)^{-1} -  W_{k^*}^A \| \nonumber \\
&\leq &  \lim_{t \to 0}  \|  \tilde{W}_{k^*}(t) \| \|  W_{k^*-1}(t)^{-1} - (W_{k^*-1}^A)^{-1} \| + \| \tilde{W}_{k^*}(t)  - \tilde{W}_{k^*}(0) \| \| (W_{k^*-1}^A)^{-1} \| \nonumber \\
&=& 0~,
\end{eqnarray}
since $W_{k^*-1}(t)^{-1}$  and $\tilde{W}_{k^*}(t)$ are both continuous at $t=0$.
Analogously we have 
$\lim_{t \to 1} W_{k^*}(t) = W_{k^*}^B $.


Finally, if either $\text{rank}(W_{k^*-1}^A) < m$ or $\text{rank}(W_{k^*-1}^B) < m$, 
we denote by $P_A$ (resp $P_B$) the orthogonal complement of $\text{span}(W_{k^*-1}^A)$ 
(resp  $\text{span}(W_{k^*-1}^B)$), and by $Q_A$ (resp $Q_B$) the orthogonal complement of 
$\text{Null}(W_{k^*}^A)$ 
(resp  $\text{Null}(W_{k^*}^B)$).
 Observe that if either $Q_A$ intersects with $P_A$ (resp  $Q_B$ intersects with $P_B$), 
 we can shrink $W_{k^*}^A$ in the intersection with no effect in the loss. 
 We can thus assume without loss of generality that $P_A \cap Q_A = \emptyset$. 
 In that case, increasing the range of $W_{k*-1}^A$ until it has rank $m$ has no effect 
 in the loss either, since the new directions will fall in the kernel of $W_{k^*}^A$. 
Therefore, by applying the necessary corrections to $W_{k*-1}^A$ and $W_{k*}^A$
 (resp $W_{k*-1}^B$ and $W_{k*}^B$) we can reduce ourselves to the previous case.

%
%
%

%

Finally, let us prove that the result is also true when $K=2$ and $\kappa>0$.
We construct the path using the variational properties of atomic norms \cite{bach2013convex}. 
When we pick the ridge regression regularization, the corresponding atomic norm is the 
nuclear norm:
$$\| X \|_{*} = \min_{UV^T = X} \frac{1}{2}( \| U \|^2 + \| V \|^2)~.$$
The path is constructed by exploiting the convexity of the variational norm $\| X\|_{*}$. 
Let $\theta^A = (W_1^A, W_2^A)$ and $\theta^B=(W_1^B, W_2^B)$, and we 
define $\tilde{W} = W_1 W_2$. Since $\tilde{W}^{\{A,B\}} = W_1^{\{A,B\}} W_2^{\{A,B\}} $, 
it results that
\begin{equation}
\label{colcol}
\| \tilde{W}^{\{A,B\}} \|_* \leq \frac{1}{2} ( \| W_1^{\{A,B\}}\|^2 + \| W_2^{\{A,B\}}\|^2)~.
\end{equation}
From (\ref{colcol}) it results that the loss $\Forr(W_1, W_2)$ can be minored by another loss 
 expressed in terms of $\tilde{W}$ of the form 
$$\E \{ | Y - \tilde{W} X |^2 \} + 2\kappa \| \tilde{W} \|_*~,$$ 
which is convex with respect to $\tilde{W}$. Thus a linear path in $\tilde{W}$ from 
$\tilde{W}^A$ to $\tilde{W}^B$ is guaranteed to be below $\Forr(\theta^{\{A,B\}})$. 
Let us define 
$$\forall~t~,~W_{1}(t), W_{2}(t) = \arg\min_{UV^T= \tilde{W}(t)} (\| U\|^2 + \|V \|^2)~. $$ 
One can verify that we can first consider a path $(\beta^A_1(s), \beta^A_2(s))$ 
from $(W_1^{A}, W_2^{A})$ to $(W_{1}(0), W_{2}(0)$ such that 
$$\forall~s~\beta_1(s) \beta_2(s) = \tilde{W}^A \text{ and } \| \beta_1(s) \|^2 + \| \beta_2(s) \|^2 \text{ decreases} ~,$$
and similarly for $(W_1^{B}, W_2^{B})$ to $(W_{1}(1), W_{2}(1)$.
The path $(\beta_{\{1,2\}}^A(s), W_{\{1,2\}}(t), \beta_{\{1,2\}}^B(s))$ satisfies (i-iii) by definition. We also verify that 
\begin{eqnarray*}
\| W_{1}(t) \|^2 + \| W_{2}(t) \|^2 &=& 2 \| \tilde{W}(t)\|_{*}   \\
&\leq & 2 (1-t) \|  \tilde{W}(0)\|_{*} + 2 t  \|  \tilde{W}(1)\|_{*} \\ 
&\leq & (1-t) ( \| W \|_{1}^2(0) + \| W \|_{2}^2(0)  ) + t (\| W \|_{1}^2(1) + \| W \|_{2}^2(1))~.
\end{eqnarray*}
Finally, we verify that the paths we have just created, when applied to $\theta^A$ arbitrary and 
$\theta^B = \theta^*$ a global minimum, are strictly decreasing, again by induction. 
For $K=1$, this is again an immediate consequence of convexity. 
For $K>1$, our inductive construction guarantees that for any $0<t<1$, 
the path $\theta(t) = (W_k(t))_{k \leq K}$ satisfies $\Forr(\theta(t)) < \Forr(\theta^A)$. 
 This concludes the proof $\square$.

\subsection{Proof of Proposition \ref{localdistprop}}

Let 
$$A(w_1, w_2) = \{ x \in \R^n; \, \langle x, w_1 \rangle \geq 0\,,\, \langle x, w_2 \rangle \geq 0\}~.$$
By definition, we have 
\begin{eqnarray}
\label{col1}
\langle w_1, w_2 \rangle_Z &=& \E \{ \max(0, \langle X, w_1 \rangle ) \max(0, \langle X, w_2 \rangle ) \} \\
&=& \int_{A(w_1, w_2)} \langle x, w_1 \rangle  \langle x, w_2 \rangle dP(x)~, \\
&=& \int_{Q({A}(w_1, w_2))}  \langle Q(x), w_1 \rangle  \langle Q(x), w_2 \rangle (d\bar{P}(Q(x)))~,  
\end{eqnarray}
where $Q$ is the orthogonal projection onto the space spanned by $w_1$ and $w_2$ and
 $d\bar{P}(x)=d\bar{P}(x_1, x_2)$ is the marginal density on that subspace. 
 Since this projection does not interfere with the rest of the proof, we abuse notation by dropping the $Q$ and still referring to $dP(x)$ as the probability density.

Now, let $r = \frac{1}{2}\| w_1 + w_2 \| = \frac{1 + \cos(\alpha)}{2}$ and $d = \frac{w_2 - w_1}{2}$.
By construction we have 
$$w_1 = r w_m - d~,~ w_2 = r w_m + d~,$$
and thus 
\begin{equation}
\label{col2}
\langle x, w_1 \rangle  \langle x, w_2 \rangle = r^2 | \langle x, w_m \rangle |^2 - | \langle x, d \rangle |^2~.
\end{equation}
By denoting $C(w_m) = \{ x \in \R^n;\, \langle x, w_m \rangle \geq 0 \}$, 
observe that $A(w_1, w_2 ) \subseteq C(w_m)$. Let us denote by $B = C(w_m) \setminus A(w_1, w_2) $ the disjoint complement. It results that 
\begin{alignat}{3}
\label{col5}
\langle w_1, w_2 \rangle_Z &= \int_{A(w_1, w_2)} &&\langle x, w_1 \rangle  \langle x, w_2 \rangle dP(x) \nonumber \\
&= \int_{C(w_m)} &&[r^2 | \langle x, w_m \rangle |^2 - | \langle x, d \rangle |^2 ] dP(x) - \nonumber \\
& &&r^2 \int_B  | \langle x, w_m \rangle |^2 dP(x) + \int_B  | \langle x, d \rangle |^2  dP(x) \nonumber \\ 
&= &&r^2 \| w_m \|_Z^2 - \underbrace{ r^2 \int_B  | \langle x, w_m \rangle |^2 dP(x)}_{E_1} - \underbrace{\int_{A(w_1, w_2)} | \langle x, d \rangle |^2  dP(x) }_{E_2}~.
\end{alignat} 
We conclude by bounding each error term $E_1$ and $E_2$ separately:
\begin{equation}
\label{col3}
0 \leq E_1 \leq r^2 |\sin(\alpha)|^2 \int_B \| x \|^2 dP(x) \leq r^2 |\sin(\alpha)|^2 2 \| \Sigma_X\|~,
\end{equation}
since every point in $B$ by definition has angle greater than $\pi/2 - \alpha$ from $w_m$. Also,
\begin{equation}
\label{col4}
0 \leq E_2 \leq \|d \|^2 \int_{A(w_1, w_2)} \| x \|^2 dP(x) \leq \frac{1 - \cos(\alpha)}{2} 2 \| \Sigma_X \|
\end{equation}
by direct application of Cauchy-Schwartz. The proof is completed by plugging the bounds from (\ref{col3}) and (\ref{col4}) into (\ref{col5})  $\square$.

\subsection{Proof of Theorem \ref{maintheo}}

Consider a generic $\alpha$ and $l \leq m$. A path from $\theta^A$ to $\theta^B$ will be constructed 
by concatenating the following paths:
\begin{enumerate}
\item from $\theta^A$ to $\theta_{lA}$, the 
best linear predictor using the same first layer as $\theta^A$, 
\item from $\theta_{lA}$ to $\theta_{sA}$, the best $(m-l)$-term approximation using perturbed 
atoms from $\theta^A$,
\item from $\theta_{sA}$ to $\theta^*$ the oracle $l$ term approximation,  
\item from $\theta^*$ to $\theta_{sB}$, the best $(m-l)$-term approximation using perturbed 
atoms from $\theta^B$,
\item from $\theta_{sB}$ to $\theta_{lB}$, the 
best linear predictor using the same first layer as $\theta^B$, 
\item from $\theta_{lB}$ to $\theta^{B}$.
\end{enumerate}
The proof will study the increase in the loss along each subpath and aggregate 
the resulting increase into a common bound. 

Subpaths (1) and (6) only involve changing the parameters of the second layer 
while leaving the first-layer weights fixed, which define a convex loss. Therefore a linear path is sufficient to guarantee that 
the loss along that path will be upper bounded by $\lambda$ on the first end 
and $\delta_{W_1^A}(m,0,m)$ on the other end. 

Concerning subpaths (3) and (4), we notice that they can also be constructed using only parameters of the second layer, 
by observing that one can fit into a single $n \times m$ parameter matrix both the 
$(m-l)$-term approximation and the oracle $l$-term approximation. 
Indeed, let us describe subpath (3) in detail ( subpath (4) is constructed analogously by replacing the role of $\theta_{sA}$ 
with $\theta_{sB}$). Let $\tilde{W}_A$ the first-layer parameter matrix associated with the 
$m-l$-sparse solution $\theta_{sA}$, and let $\gamma_A$ denote its second layer coefficients, which is a $m$-dimensional vector
with at most $m-l$ non-zero coefficients. 
Let $W_{*}$ be the first-layer matrix of the $l$-term oracle approximation, and $\gamma_{*}$ the corresponding second-layer coefficients. 
Since there are only $m-l$ columns of $\tilde{W}_A$ that are used, corresponding to the support of $\gamma_A$, we can 
consider a path $\bar{\theta}$ that replaces the remaining $l$ columns with those from $W_{*}$ while keeping the second-layer vector $\gamma_A$ fixed. Since the modified columns correspond to zeros in $\gamma_A$, such paths have constant loss. 
Call $\bar{W}$ the resulting first-layer matrix, containing both the active $m-l$ active columns of $\tilde{W}_A$ and the $l$ columns of $W_{*}$ in the positions determined by the zeros of $\gamma_A$. 
Now we can consider the linear subpath that interpolates between $\gamma_A$ and $\gamma_{*}$ while keeping the first layer fixed at $\bar{W}$. 
Since again this is a linear subpath that only moves second-layer coefficients, it is non-increasing thanks to the convexity of the loss while fixing the first layer. We easily verify that at the end of this linear subpath we are using the oracle $l$-term approximation, which has loss $e(l)$, and therefore 
subpath (3) incurs in a loss that is bounded by its extremal values $\delta_{W_1^A}(m-l, \alpha, m)$ and $e(l)$.

Finally, we need to show how to construct the subpaths (2) and (5), which are the most delicate step since they cannot be bounded using
convexity arguments as above. 
Let $\tilde{W}_A$ be the resulting perturbed first-layer parameter matrix 
with $m-l$ sparse coefficients $\gamma_A$.
Let us consider an auxiliary regression of the form 
$$\overline{W} = [ W^A ; \tilde{W}_A] ~\in \R^{n \times 2m}~.$$
and regression parameters 
$$\overline{\beta}_1 = [ \beta_1; 0]~,~\overline{\beta}_2 = [0; \gamma_A]~.$$
Clearly 
$$\E\{ | Y - \overline{\beta}_1 \overline{W} |^2 \} + \kappa \| \overline{\beta}_1 \|_1 = \E\{ | Y - \beta_1 W^A |^2 \} + \kappa \| {\beta}_1 \|_1 $$ 
 and similarly for $\overline{\beta}_2$. By convexity, the augmented linear path $\eta(t) =(1- t) \overline{\beta}_1 + t \overline{\beta}_2$ thus satisfies 
$$\forall~t~,\overline{L}(t) = \E\{ | Y - \eta(t) \overline{W} |^2 \} + \kappa \| \eta(t) \|_1 \leq \max(\overline{L}(0), \overline{L}(1))~. $$
Let us now approximate this augmented linear path with a path in terms of first and second layer weights. 
We consider
$$\eta_1(t) = (1-t) W^A + t \tilde{W}_A~,\text{ and}~\eta_2(t) = (1- t) {\beta}_1 + t \gamma_A~.$$
We have that 
\begin{alignat}{3}
\label{bub1}
\Forr(\{ \eta_1(t), \eta_2(t) \}) &= &&\ \E \{ | Y - \eta_2(t) Z(\eta_1(t) ) |^2 \} + \kappa \| \eta_2(t) \|_1  \\ 
&\leq  &&\ \E \{ | Y - \eta_2(t) Z(\eta_1(t) ) |^2 \} + \kappa(  ( 1-t) \| {\beta}_1\|_1 + t \| \gamma_A \|_1 ) \nonumber \\
& = &&\ \overline{L}(t) + \E \{ | Y - \eta_2(t) Z(\eta_1(t) ) |^2 \} \nonumber \\
& - &&\ \E \{ | Y - (1-t) \beta_1 Z(W^A) - t \gamma_A Z(\tilde{W}_A) |^2 \} ~.
\end{alignat}
Finally, we verify that
{\small 
\begin{eqnarray}
\label{bub2}
& \left | \E \{ | Y - \eta_2(t) Z(\eta_1(t) ) |^2 \}  - \E \{ | Y - (1-t) \beta_1 Z(W^A) - t \gamma_A Z(\tilde{W}_A) |^2 \} \right|  \leq\\
& \leq 4  \alpha \max(\E |Y|^2, \sqrt{\E|Y^2|}) \| \Sigma_X \| ( \kappa^{-1/2} + \alpha \sqrt{\E|Y^2|} \kappa^{-1}) + o(\alpha^2)~. \nonumber
\end{eqnarray}}
Indeed, from Proposition \ref{localdistprop}, and using the fact that 
$$\forall~i\leq M,\, t \in [0,1]~,~\left| \angle( (1-t)w^A_i + t \tilde{w}^A_i ; w^A_i) \right| \leq \alpha~,~ \left| \angle( (1-t)w^A_i + t \tilde{w}^A_i ; \tilde{w}^A_i) \right| \leq \alpha $$
we can write 
$$(1-t) \beta_{1,i} z(w^A_i) + t \gamma_{A,i} z(\tilde{w}^A_i) \stackrel{d}{=} \eta_2(t)_i z(\eta_1(t)_i) + n_i ~,$$
with $\E\{ |n_i |^2 \} \leq 4 |\eta_2(t)_i|^2 \| \Sigma_X \| \alpha^2 + O(\alpha^4)~$ and $\E |n_i| \leq 2 |\eta_2(t)_i| \alpha \sqrt{\| \Sigma_X\|}$ using concavity of the moments.
Thus 
\begin{eqnarray*}
&& \left | \E \{ | Y - \eta_2(t) Z(\eta_1(t) ) |^2 \}  - \E \{ | Y - (1-t) \beta_1 Z(W^A) - t \gamma_A Z(\tilde{W}_A) |^2 \} \right| \\
 &\leq& 2\E \left\{  \sum_i (Y - \eta_2(t) Z(\eta_1(t) )) n_i  \right\} + \E \left\{ | \sum_i n_i |^2 \right\} \\
 &\leq & 4\left(\alpha \sqrt{\E|Y^2|} \| \Sigma_X\|  \| \eta_2 \| + \alpha^2 (\| \eta_2 \|_1)^2  \| \Sigma_X \| \right) \\
 &\leq & 4 \alpha \max(1, \sqrt{\E|Y^2|}) \| \Sigma_X \| ( \| \eta_2 \|_1 + \alpha \| \eta_2 \|_1^2) + o(\alpha^2) \\
 & \leq & 4 \alpha \max(\sqrt{\E|Y^2|}, {\E|Y^2|}) \| \Sigma_X \|  ( \kappa^{-1} + \alpha \sqrt{\E|Y^2|} \kappa^{-2}) + o(\alpha^2)~,
\end{eqnarray*}
 which proves (\ref{bub2}). 
 
We have just constructed a path from $\theta^A$ to $\theta^B$, in which all subpaths except (2) and (5) have energy maximized at the extrema due to convexity, given respectively by $\lambda$, $\delta_{W_A^1}(m, 0, m)$, $\delta_{W_A^1}(m-l, \alpha, m)$, $e(l)$, $\delta_{W_B^1}(m-l, \alpha, m)$, and 
$\delta_{W_B^1}(m, 0, m)$.  For the two subpaths (2) and (5), (\ref{bub2}) shows that it is sufficient to add the corresponding upper bound to the linear subpath, which is of the form $C \alpha + o(\alpha^2)$ where $C$ is an explicit constant independent of $\theta$. Since $l$ and $\alpha$ are arbitrary, we are free to pick the infimum, which concludes the proof. $\square$
 


\subsection{Proof of Corollary \ref{maincoro}}

Let us consider a generic first layer weight matrix $W \in \R^{n \times m}$. Without loss of generality, we can assume that $\| w_k \|=1$ for all $k$, since increasing the norm of $\|w_k\|$ 
within the unit ball has no penalty in the loss, and we can compensate this scaling in the second layer
thanks to the homogeneity of the half-rectification. Since this results in an attenuation of these second layer weights, 
they too are guaranteed not to increase the loss. 

From \cite{vershynin2010introduction} [Lemma 5.2] we verify that the covering number $\mathcal{N}(S^{n-1}, \epsilon)$ of the Euclidean unit sphere $S^{n-1}$ 
satisfies 
$$\mathcal{N}(S^{n-1}, \epsilon) \leq \left( 1 + \frac{2}{\epsilon} \right)^n~,$$
which means that we can cover the unit sphere with an $\epsilon$-net of size $\mathcal{N}(S^{n-1}, \epsilon) $. 

Let $0< \eta < n^{-1} ( 1 + n^{-1})^{-1}$, and let us pick, for each $m$, $\epsilon_m = m^{\frac{\eta-1}{n} }$. 
Let us consider its corresponding $\epsilon$-net of size 
$$u_m = \mathcal{N}(S^{n-1}, \epsilon_m) \simeq \left( 1 + \frac{2}{\epsilon_m} \right)^n \simeq m^{1-\eta} ~.$$
Since we have $m$ vectors in the unit sphere, it results from the pigeonhole principle that at least one 
element of the net will be associated with at least $v_m = m u_m^{-1} \simeq m^{\eta}$ vectors; in other words, 
we are guaranteed to find amongst our weight vector $W$ a collection $Q_m$ of $v_m \simeq m^\eta$ vectors that are all 
at an angle at most $2\epsilon_m$ apart. 
Let us now apply Theorem \ref{maintheo} by picking $n=v_m$ and $\alpha = \epsilon_m$. 
We need to see that the terms involved in the bound all converge to $0$ as $m \to \infty$. 

The contribution of the oracle error $e(v_m) - e(m)$ goes to zero as $m\to \infty$ by the 
fact that $\lim_{m \to \infty} e(m)$ exists (it is a decreasing, positive sequence) and that $v_m \to \infty$.

Let us now verify that $\delta(m - v_m, \epsilon_m, m)$ also converges to zero.
We are going to prune the first layer by removing one by one the vectors in $Q_m$.
Removing one of these vectors at a time incurs in an error of the order of $\epsilon_m$. 
Indeed, let $w_k$ be one of such vectors and let $\beta'$ be the solution of 
$$\min_{\beta'} E(\beta') = \min_{\beta'=(\beta_f;\beta_k) \in \R^{k}} \E \{ |Y - \beta_f^T Z(W_{-k}) - \beta_k z(w_k)|^2 \} + \kappa ( \| \beta_f \|_1 + |\beta_k| )~, $$
where $W_{-k}$ is a shorthand for the matrix containing the rest of the vectors that have not been discarded yet.
  Removing the vector $w_k$ from the first layer
 increases the loss by a factor that is upper bounded by $E(\beta_p) - E(\beta)$, where 
$$(\beta_p)_j = \left\{
\begin{array}{rl}
 \beta'_j  & \text{ for  } j < k -1 ~,\\
 \beta'_{k-1} + \beta'_{k } & \text{ otherwise.}
\end{array}\right.~, $$
 since now $\beta_p$ is a feasible solution for the pruned first layer.

Let us finally bound $E(\beta_p) - E(\beta)$. 

Since $\angle(w_k,w_{k-1}) \leq \epsilon_m $, 
it results from Proposition \ref{localdistprop} that 
$$z(w_k) \stackrel{d}{=}  z(w_{k-1}) + n~,$$
with $\E\{|n|^2\} \leq C\alpha^2$ for some constant $C$ independent of $m$. 
By redefining $p_1 = Y - \beta_p^T Z(W_{-k}) - \frac{1}{2} n$ and 
$p_2 = \frac{1}{2}n$, we have 
\begin{alignat*}{3}
&\E \{ | Y - \beta_p^T Z(W_{-k}) |^2 \} - \E \{ | Y - {\beta'}^T Z(W_{-k}) - \beta_k z(w_k) |^2 \} \\
=\ &\E\{ | p_1 + p_2 |^2 \} - \E \{ |p_1 - p_2 |^2 \} \\
=\ &4\E \{ |p_1 p_2| \} \\
\leq\ &\sqrt{ \E\left\{ \left|Y - \beta_p^T Z(W_{-k}) - \frac{1}{2} n \right|^2 \right\} } \sqrt{ \E \{ |n|^2 \} } \\
\leq\  &(C+\alpha) \alpha \simeq \epsilon_m~,
\end{alignat*}
where $C$ only depends on $\E\{|Y|^2\}$. We also verify that $\| \beta_p\|_1 \leq \| \beta'\|_1$.

It results that removing $|Q_m|$ of such vectors incurs an increase of the loss at most 
$|Q_m| \epsilon_m \simeq m^\eta m^{\frac{\eta-1}{n} } = m^{\eta + \frac{\eta-1}{n}}$. 
Since we picked $\eta$ such that $\eta + \frac{\eta-1}{n} <0$, this term converges to zero. The proof is finished. $\square$

%
%
%
%
%

%% file: visualization.tex
\section{Cartoon of Algorithm}
\label{AlgCartoon}

\begin{figure}
\begin{center}
\scalebox{1}{\includegraphics[width=1.0\textwidth]{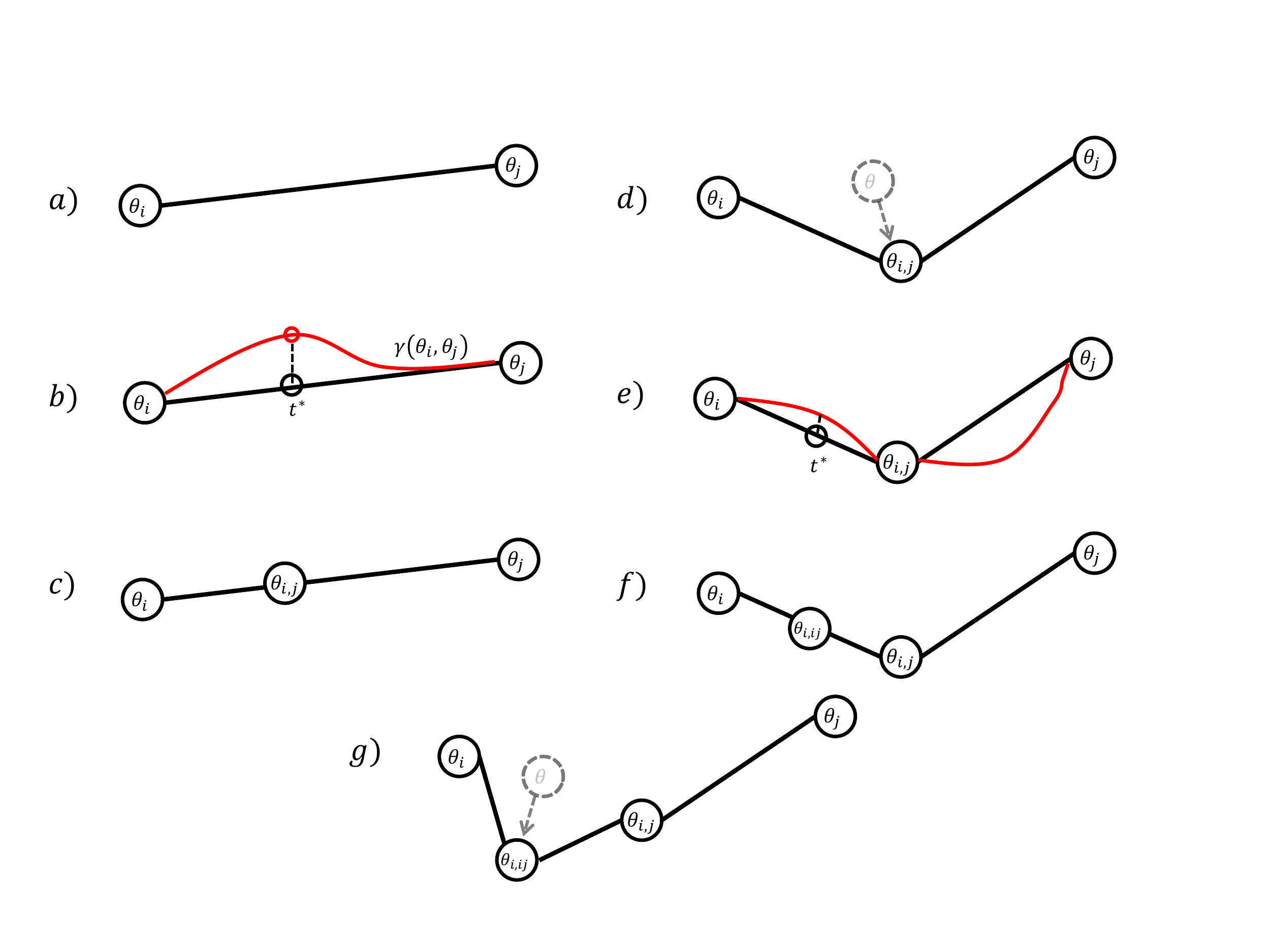}}
\end{center}
\caption{A cartoon of the algorithm.  $a):$ The initial two models with approximately the same loss, $L_0$. $b):$ The interpolated loss curve, in red, and its global maximum, occuring at $t=t^*$. $c):$ The interpolated model $\Theta(\theta_i, \theta_j, t^*)$ is added and labeled $\theta_{i,j}$.  $d):$ Stochastic gradient descent is performed on the interpolated model until its loss is below $\alpha L_0$. $e):$ New interpolated loss curves are calculated between the models, pairwise on a chain.  $f):$ As in step $c)$, a new model is inserted at the maxima of the interpolated loss curve between $\theta_i$ and $\theta_{i,j}$.  $g):$  As in step $d)$, gradient descent is performed until the model has low enough loss.}
\label{fig:cartoon}
\end{figure}

Refer to Fig. \ref{fig:cartoon}.

\section{Visualization of Connection}
\label{visualization}

\begin{figure}
\centering
\includegraphics[width=.4\textwidth]{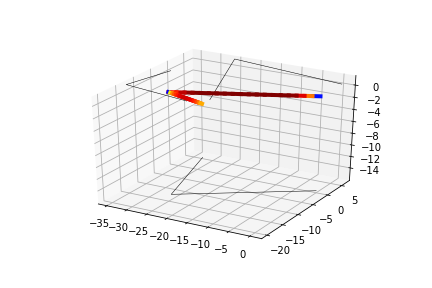}
\includegraphics[width=.4\textwidth]{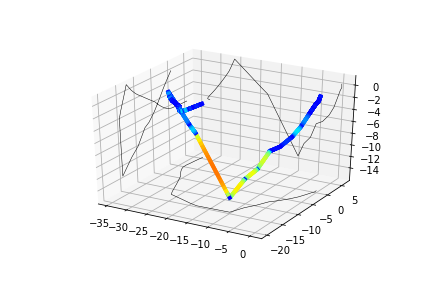}
\includegraphics[width=.4\textwidth]{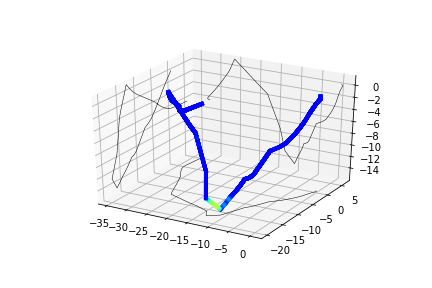}
\includegraphics[width=.4\textwidth]{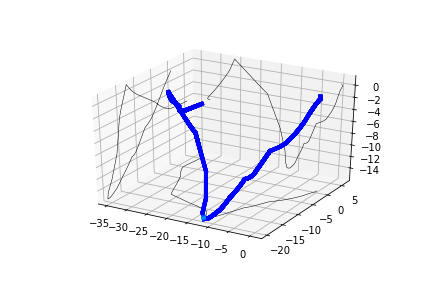}
\caption{Snapshots of Dynamic String Sampling in action for the quadratic regression task.  The string's coordinates are its projections onto the three most important principal axes of the fully converged string.  (Top Left) One step into the algorithm, note the high loss between all of the vertices of the path. (Top Right) An intermediate step of the algorithm.  Portions of the string have converged, but there are still regions with high interpolated loss. (Bottom Left) Near the end of the algorithm.  Almost the entire string has converged to low loss.  (Bottom Right) The algorithm has finished.  A continuous path between the models has been found with low loss.} 
\label{connfigs}
\end{figure}

 Because the weight matrices are anywhere from high to extremely high dimensional, for the purposes of visualization we projected the models on the connecting path into a three dimensionsal subspace.  Snapshots of the algorithm in progress for the quadratic regression task are indicated in Fig. \ref{connfigs}.  This was done by vectorizing all of the weight matrices for all the beads for a given connecting path, and then performing principal component analysis to find the three highest weight projections for the collection of models that define the endpoints of segments for a connecting path---i.e., the $\theta_i$ discussed in the algorithm.  We then projected the connecting string of models onto these three directions.  
 
 The color of the strings was chosen to be representative of the test loss under a log mapping, so that extremely high test loss mapped to red, whereas test loss near the threshold mapped to blue.  An animation of the connecting path can be seen on our \href{github.com/danielfreeman11/convex-nets/blob/master/Writeup/Plots/quadratic.pathinterp.errorvis.gif}{Github page}.
 
 Finally, projections onto pairs of principal components are indicated by the black curves.
 
\section{A Disconnection}
\label{sec:disconnect}

\subsection{A Disconnection}
\label{symdisc}

\begin{figure}
\centering
\includegraphics[width=.4\textwidth]{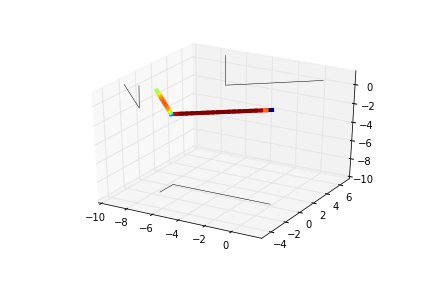}
\includegraphics[width=.4\textwidth]{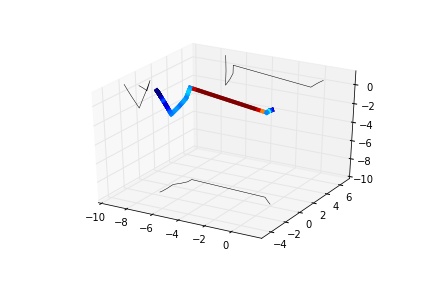}
\includegraphics[width=.4\textwidth]{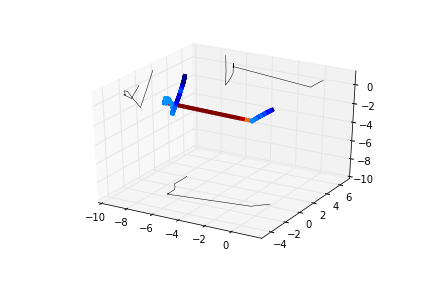}
\caption{These three figures are projections of the components of the 12-dimensional weight matrices which comprise the models on the string produced by the DSS algorithm.  The axes are the principal components of the weight matrices, and the colors indicate test error for the model.  For more details on the figure generation, see Appendix \ref{visualization}. (Left) The string of models after 1 step.  Note the high error at all points except the middle and the endpoints.  (Middle) An intermediate stage of the algorithm.  Part of the string has converged, but a persistent high-error segment still exists.  (Right) Even after running for many steps, the error persists, and the algorithm does not converge.}
\label{discfigs}
\end{figure}

 As a sanity check for the algorithm, we also applied it to a problem for which we know that it is not possible to connect models of equivalent power by the arguments of section \ref{disconnect}.  The input data is 3 points in $\mathbb{R}^2$, and the task is to permute the datapoints, i.e. map $\{x_1,x_2,x_3\} \to \{x_2,x_3,x_1\}$.  This map requires at least 12 parameters in general for the three linear maps which take $x_i\to x_j$ for $i,j \in \{\{1,2\},\{2,3\},\{3,1\}\}$.  Our archticture was a 2-3-2 fully connected neural network with a single relu nonlinearity after the hidden layer---a model which clearly has 12 free parameters by construction.  The two models we tried to connect were a single model, $\theta$, and a copy of $\theta$ with the first two neurons in the hidden layer permuted, $\tilde{\theta_{\sigma}}$.  The algorithm fails to converge when initialized with these two models.  We provide a visualization of the string of models produced by the algorithm in Fig. \ref{discfigs}.
 
 In general, a persistent high interpolated loss between two neighboring beads on the string of models could arise from either a slowly converging, connected pair of models or from a truly disconnected pair of models.  ``Proving'' a disconnection at the level of numerical experiments is intractable in general, but a collection of negative results---i.e., failures to converge---are highly suggestive of a true disconnection.
 
